\documentclass[11pt,a4paper]{article}

\usepackage[margin=1in]{geometry}
\usepackage{setspace}
\usepackage[T1]{fontenc}
\usepackage[utf8]{inputenc}
\usepackage{lmodern}

\usepackage{xcolor}  
\usepackage{listings}
\usepackage{courier}  
\usepackage{microtype}
\usepackage{graphicx}
\usepackage{subcaption}
\usepackage[most]{tcolorbox}

\newtcolorbox{agentprompt}{%
  colback=gray!5,%
  colframe=gray!50,%
  boxrule=0.5pt,%
  left=10pt,%
  right=10pt,%
  top=8pt,%
  bottom=8pt,%
  arc=2pt,%
  fontupper=\small\itshape%
}
\usepackage{booktabs}
\usepackage{siunitx}
\usepackage{amsmath}
\usepackage{hyperref}
\hypersetup{
	colorlinks=true,
	linkcolor=blue,
	citecolor=blue,
	urlcolor=blue
}

\usepackage{authblk}

\usepackage{titlesec}
\titleformat{\paragraph}[block]{\normalfont\normalsize\bfseries}{\theparagraph}{1em}{}

\usepackage[round,authoryear]{natbib}

\title{Context is all you need: Towards autonomous model-based process design using agentic AI in flowsheet simulations}
\author[1,*]{Pascal Schäfer}
\author[1,2]{Lukas J. Krinke}
\author[1]{Martin Wlotzka}
\author[1]{Norbert Asprion}
\affil[1]{BASF SE, Ludwigshafen, Germany}
\affil[2]{RWTH Aachen University, Aachen, Germany}
\affil[*]{Corresponding author: \texttt{pascal.schaefer@basf.com}}

\date{\today}

\begin{document}

\maketitle

\begin{abstract}
Agentic AI systems integrating large language models (LLMs) with reasoning and tool-use capabilities are transforming software development, yet their application in chemical process flowsheet modelling remains largely unexplored.
 We present an agentic AI framework that provides assistance in an industrial flowsheet simulation environment.
 We show that GitHub Copilot~\citep{githubCopilot}, using state-of-the-art LLMs such as Claude Opus 4.6~\citep{anthropicClaude}, can generate valid syntax for our in-house process modelling tool Chemasim from technical documentation and a few commented examples.
 Building on this, we develop a multi-agent system that decomposes process development tasks: one agent solves the abstract engineering problem while another implements the solution as Chemasim code.
 We demonstrate the framework on typical flowsheet modelling examples, including a reaction/separation process, a pressure-swing distillation, and a heteroazeotropic distillation with entrainer selection, and discuss limitations and future directions.

\end{abstract}
\noindent\textbf{Topical heading:} Process Systems Engineering


\section{Introduction}

The chemical industry is currently facing major challenges with one particular hurdle lying in the adaptation of chemical processes to new feedstocks such as biomass or CO$_2$ (e.g., \citep{Sheldon2014,Artz2018}).
 In this context, process flowsheet simulation, being the evergreen in the chemical engineer's digital tool case for more than 50 years (e.g., \cite{Sargent1964,Hegner1973}), plays a crucial role as it can provide decision support on which process alternatives to discard or pursue further with reduced need for costly and time-consuming experiments.
	
Classically, the synthesis of chemical processes is conducted in an iterative yet mostly sequential manner where a conceptual flowsheet is developed based on heuristics (with the most prominent ones being certainly the hierarchical procedure from \cite{Douglas1985}) and best practices followed by a simulation using the tool of choice.
 If necessary, the conceptual flowsheet is then refined based on the simulation results and the procedure is repeated until a final process design is obtained.
 With the advent of numerical optimization techniques starting in the 1980s (cf. \cite{Biegler2004}), researchers have started developing more systematic approaches for obtaining optimal process designs.
 In particular, the ability to find suitable values for remaining degrees of freedom in a structurally fixed flowsheet using mathematical programming is well established in the literature (e.g., \cite{Dowling2015}) and nowadays sometimes already available in industrially applied flowsheet simulators.
 In contrast, finding the optimal process structure remains much more challenging.
 Here, the majority of academic research focuses on superstructure-based approaches (e.g., \cite{Mencarelli2020}). However, despite a few trials in open-source frameworks (e.g., \cite{Chen2018}), they have not found their way into industrially applied flowsheet simulators.

In recent years, reinforcement learning (RL) has emerged as an alternative that omits the need for a superstructure and avoids transferring the model to a different modelling environment, thereby allowing for systematic decision support for process synthesis tasks directly within the simulation environment of choice.
 In these works, an RL agent is trained to modify a flowsheet by executing a sequence of possible actions (e.g., adding/removing units, redirecting streams, etc.) aiming to maximize a so-called reward function representing the process performance (e.g., \cite{Gao2024}).
 Despite the very promising fact that these approaches work essentially independently of the underlying flowsheet simulator, they are - to the best of our knowledge - currently limited to simplified process models and non-complex and pre-defined thermodynamic systems.
 Moreover, the generalization capabilities of trained agents from one synthesis task to another remain to be demonstrated.
 In fact, most works focus on one distinct process synthesis task, usually a reaction step with a subsequent separation sequence \citep{Gottl2021,Stops2023,Gao2025}.
 Broadening this view, \cite{Gottl2025} simultaneously considers multiple types of mixtures to be separated during training with the agent afterwards being able to cope with arbitrary feed compositions for each separation task.
 Nevertheless, the generalization to unseen mixtures, e.g., based on comparable thermodynamic properties, remains an open question.

Recognizing conceptual similarities between chemical process modelling and software development, using large language models (LLMs) - particularly, in the form of agentic AI systems that integrate them with reasoning and tool-use capabilities - could provide a promising alternative to the above-mentioned approaches.
 This then gives rise to three primary questions:
\begin{enumerate}
	\item Can LLMs generate valid syntax for implementing these solutions within an industrial process modelling tool?
	\item Can agentic frameworks effectively handle the process simulation engine to finally get converged solutions?
	\item Are LLMs capable of solving abstract chemical process modelling problems, e.g., synthesizing separation sequences for given feed mixtures, by interpreting the thermodynamic system?
\end{enumerate}

Allowing for direct textual editing of input files (that can alternatively be generated using a graphical user interface), BASF's in-house process modelling tool Chemasim seems particularly well suited for such an approach.
 Note that in other cases, standardized textual representations of process flowsheet models (e.g., SFILES 2.0 \citep{Vogel2023} or DEXPI process \citep{Cameron2024}) exist and would represent an alternative intermediate format to act on, while a pre-defined interface to the simulation software would translate them into actual changes in the flowsheet model.
 
In contrast to assisting with code snippets for general-purpose programming languages (e.g., Python, C++, etc.), we face the challenge that the underlying LLMs are usually not well-trained on a highly domain-specific and individual syntax due to a lack of training data.
 This is even more true for Chemasim, an in-house tool with very limited public exposure in the form of an attempted open-source version several years ago \citep{Hasse2006}.

LLMs have nevertheless been shown to successfully learn domain-specific languages in the context of chemistry and chemical engineering.
 Examples include creating solver input files and meshes for computational fluid dynamics \citep{Pandey2025}, generating input files for quantum chemistry software \citep{Jacobs2025} and molecular dynamics simulations \citep{Shi2025}.
 For the latter, there already exists an agentic framework capable of autonomously running simulations and analyzing the results \citep{Mendible2025}.
 Closer to our scope, \cite{Rupprecht2025} used LLMs to generate Modelica~\citep{modelica} syntax for modelling of the transient behavior of reactors, i.e., to freely write the individual equations of the differential-algebraic equation system.
 There, the authors compare different LLMs, including a fine-tuned version of the open-source LLM Llama 3.1~\citep{metaLlama} and the proprietary GPT-4o~\citep{openai} model.
 Even though the authors demonstrate the general ability of LLMs to generate valid code, they also report substantial issues regarding semantic accuracy.
 Moreover, the benefit of fine-tuning a weaker open-source LLM cannot be seen from the results.

Regarding the use of agentic AI systems for chemical flowsheet modelling, \cite{Liang2026} let a single agent based on a more recent LLM (Claude Sonnet 4.0~\citep{anthropicClaude}) interact with the commercial flowsheet simulator AVEVA Process Simulation (APS)~\citep{avevaProcessSimulation} to solve a simple process synthesis problem.
 More precisely, the authors are led by the agent through a step-by-step dialogue to model and simulate a single distillation column for separating a binary mixture of water and methanol, giving very precise tasks to the agent.
 Tackling more abstract problems using agentic AI frameworks, \cite{Prasopsanti2026} propose a multi-agent system based on the open-source LLM Qwen 2.5~\citep{qwen} for feasibility analysis and design of separation tasks.
 Here, the agents have access to knowledge libraries, physical property databases and a variety of tools for performing thermodynamic calculations and process engineering computations.
 The authors demonstrate the capabilities on a number of case studies tackling the conceptual qualitative design of separation sequences for multi-component mixtures.

The present work focuses on the ability of the most recent generation of LLMs to conduct both conceptual process synthesis and implementation of the resulting flowsheet model in Chemasim syntax.
 To this end, we provide the LLM with context in the form of technical documentation and a few well-commented examples and do not rely on fine-tuning.
 Furthermore, we show how interfaces to the simulation software can enable the use of LLMs in an agentic AI framework, allowing it to react to simulation results and error messages and thus enabling autonomous model development.
 In particular, we develop a multi-agent system (cf. \cite{Rupprecht2026} for an overview of their application within chemical engineering).
 Therein, a process developer with access to tools for thermodynamic calculations and the ability for free coding is responsible for solving the abstract chemical process modelling problem. 
 The resulting unit-by-unit process plan is then provided to a Chemasim modelling agent for implementation using valid syntax and interaction with the simulation software.

The remainder of this paper is structured as follows: First, we introduce the Chemasim modelling language and illustrate key syntax elements required for implementing unit operations and specifications.
 Next, we describe the proposed multi-agent system, including the responsibilities of the individual agents and the workflow for generating valid Chemasim code and interacting with the simulation engine. 
 We then present results for several process development case studies and discuss the capabilities and limitations of the framework. Along these lines, we outline directions for future work.

\section{Chemasim modelling language}

Chemasim is BASF's in-house process modelling tool for equation-based steady-state and dynamic simulation of chemical processes.
 Its development started several decades ago with the earliest references dating back to the 1970s \citep{Hegner1973}. In contrast to most commercially available flowsheet simulators, Chemasim is centered around a textual flowsheet representation.
 While graphical user interfaces exist, they mainly serve to generate the underlying text file inserting valid syntax elements representing unit operations and their connections, user-defined functions, etc.

The Chemasim input file is mostly structured by the definition of unit operations.
 These then include the input and output streams as well as values for the default degrees of freedom.
 Note that commonly, Chemasim requires the users to define the thermodynamic state of the outgoing streams of each unit operation.

For example, the syntax for modelling a mixer, as the simplest unit operation, which in this case mixes two liquid streams of water and ethanol at 25°C into one output stream, is given by:

\vspace{0.5\baselineskip}
\noindent\begin{minipage}{\textwidth}
\begin{lstlisting}[language=Python, caption={Example of a Chemasim mixer unit definition.}, escapechar=@]
@\textcolor{red}{MISCHER}@: MX_01 < 25. GRAD, REIN aus BASIS, REAL aus BASIS >
  ZULAUF: STROM Water < 100 KG/H ( 1 H2O ), FLUSS 25 GRAD >
  ZULAUF: STROM Ethanol < 100 KG/H ( 1 EtOH ), FLUSS 25 GRAD >
  FLUSS:  STROM Out
\end{lstlisting}
\end{minipage}
\vspace{0.5\baselineskip}

The mixer can then also serve as a heater by adding a heat input. By default, Chemasim then expects the user to define the amount of heat.
 Alternatively, this can be set free and a different specification can be made, e.g., a specific outlet temperature as done below.
 Note that in this case, it is crucial to define the correct thermodynamic state of the outlet stream (here, a 50/50 mix of water and ethanol at 95°C is fully evaporated, which also requires the definition of a pressure for the unit) to ensure a valid simulation.

\vspace{0.5\baselineskip}
\noindent\begin{minipage}{\textwidth}
\begin{lstlisting}[language=Python, caption={Example of a Chemasim mixer unit definition modelling an evaporator.}, escapechar=@]
@\textcolor{red}{MISCHER}@: MX_01 < 25. GRAD, 1. BAR, REIN aus BASIS, REAL aus BASIS >
  ZULAUF: STROM Water < 100 KG/H ( 1 H2O ), FLUSS 25 GRAD >
  ZULAUF: STROM EtOH < 100 KG/H ( 1 EtOH ), FLUSS 25 GRAD >
  DAMPF:  STROM Out
  @\textcolor{brown}{WAERMEZULAUF}@: STROM Qheat
  @\textcolor{green}{SPEZI}@: Qheat_f < FREI WAERME STROM Qheat >
  @\textcolor{green}{SPEZI}@: Tout_s < 95 GRAD STROM Out >
\end{lstlisting}
\end{minipage}
\vspace{0.5\baselineskip}

Another example of a more complex unit operation, a distillation column, is given below. There, a large feed consisting of water and methanol being at its boiling point is separated.
 Having a total condenser and a reboiler, there are two degrees of freedom to be specified (by default, the two heat duties).
 For start-up, i.e., first simulation without initial values, we here specify the reflux ratio and the bottom mass flow rate.

\vspace{0.5\baselineskip}
\noindent\begin{minipage}{\textwidth}
\begin{lstlisting}[language=Python, caption={Example of a Chemasim distillation column unit definition.}, escapechar=@]
@\textcolor{red}{KOLONNE}@: K100 < 16 STUFEN, VERD, TOTALKOND, 25. GRAD, 1. BAR,
                ANSTARTHILFE GASANTEIL 0.5 kg/kg,
				REIN aus BASIS, REAL aus BASIS >
  ZULAUF: STROM F100, STUFE 8 < 50000. KG/H ( 18. % H2O, 82. % MeOH),
    							FLUSS 1. BAR >
  SUMPF:  STROM B100
  KOPF:   STROM D100
  @\textcolor{brown}{WAERMEABZUG}@: STROM QK_K100, KOND
  @\textcolor{brown}{WAERMEZULAUF}@: STROM QV_K100, VERD
  @\textcolor{green}{SPEZI}@: QK_K100_f < FREI WAERME STROM QK_K100 >
  @\textcolor{green}{SPEZI}@: QV_K100_f < FREI WAERME STROM QV_K100 >
  @\textcolor{green}{SPEZI}@: RR_s < RV in KG/KG 1. FLUSS APPARAT K100 >
  @\textcolor{green}{SPEZI}@: Bottom_s < 9000. KG/H MENGE STROM B100 >
\end{lstlisting}
\end{minipage}
\vspace{0.5\baselineskip}

Finally, below, we present an example of re-specification of the distillation column from above, where the specification of the reflux ratio and bottom mass flow rate is replaced by the desired purities in the distillate and bottom stream.

\vspace{0.5\baselineskip}
\noindent\begin{minipage}{\textwidth}
\begin{lstlisting}[language=Python, caption={Example for a re-specification of a distillation column in Chemasim.}, escapechar=@]
  @\textcolor{gray}{\#SPEZI: RR\_s  < RV in KG/KG 1. FLUSS APPARAT K100 >}@
  @\textcolor{gray}{\#SPEZI: Bottom\_s < 9000. KG/H MENGE STROM B100 >}@ 
  @\textcolor{green}{SPEZI}@: xD_s < 0.99 GEWANTEIL METHANOL STROM D100 >
  @\textcolor{green}{SPEZI}@: xB_s < 1E-4 GEWANTEIL METHANOL STROM B100 >
\end{lstlisting}
\end{minipage}
\vspace{0.5\baselineskip}

Besides these options for trivial specifications setting a certain quantity to a fixed value, Chemasim also allows for more complex specifications.
 In particular, the user can also set arbitrary linear combinations of quantities to a fixed value.
 Furthermore, Chemasim also offers a functional language to define more complex relationships between quantities.
 For instance, below, we present an example where the distillate-to-feed ratio of the column above is set to a certain value defined within a constant.

\vspace{0.5\baselineskip}
\noindent\begin{minipage}{\textwidth}
\begin{lstlisting}[language=Python, caption={Example specifying the distillate-to-feed ratio of column using the functional language in Chemasim.}, escapechar=@]
  @\textcolor{orange}{VARIABLE}@: mF < MENGE in KG/H STROM F100 >
  @\textcolor{orange}{VARIABLE}@: mD < MENGE in KG/H STROM D100 >
  @\textcolor{orange}{KONSTANTE}@: D2F < 0.82 >
  @\textcolor{blue}{FUNKTION}@: D2F < @\textcolor{orange}{VARIABLE}@ mD - @\textcolor{orange}{KONSTANTE}@ D2F * @\textcolor{orange}{VARIABLE}@ mF >
  @\textcolor{green}{SPEZI}@: D2F_s < 0 = @\textcolor{blue}{FUNKTION}@ D2F >
\end{lstlisting}
\end{minipage}
\vspace{0.5\baselineskip}

Further examples of Chemasim syntax can be found in the supplementary information concerning the context files provided to the agents.
 These include, among others, Chemasim implementations of the processes presented in a series of articles from \cite{Luyben2010,Luyben2011a,Luyben2011b} including the production of cumene and ethyl benzene.
 We emphasize that these implementations rather serve as examples for the Chemasim syntax of an entire process flowsheet, i.e., interconnecting multiple unit operations.
 They are not accessed as examples for process synthesis concepts suggested by the agentic AI framework.

\section{Multi-agent system}

\begin{figure}[htbp]
	\centering
	\includegraphics[width=\textwidth]{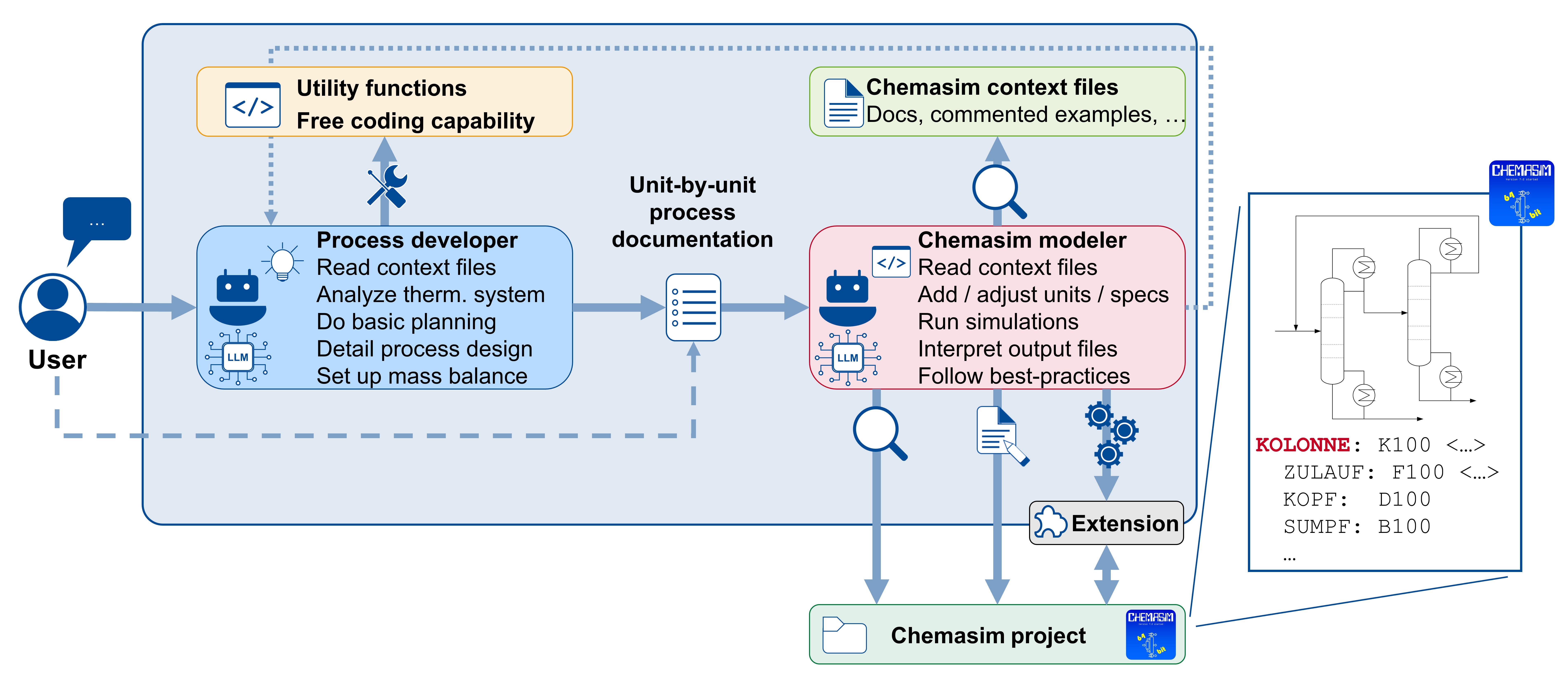}
	\caption{Multi-agent system architecture.}
	\label{fig:mas}
\end{figure}

The multi-agent system developed in this work consists of two agents based on state-of-the-art LLMs with distinct responsibilities as illustrated in Figure~\ref{fig:mas}.
 Here, the process development agent solves the abstract process synthesis task, e.g., creating a separation sequence for a given feed mixture.
 It has access to tools for performing thermodynamic computations and can freely code, i.e., write scripts in Python to perform calculations.
 A detailed unit-by-unit documentation of the process design is then provided to the Chemasim modelling agent for implementation using valid syntax.
 This agent has access to relevant context files including technical documentation, commented examples and how-tos regarding the Chemasim modelling syntax.
 Access to the Chemasim project is possible by inspecting and editing the textual input files, reading values from output files, and using an extension to perform Chemasim simulations autonomously.
 In the following, we describe the individual parts in more detail.
 
\subsection{Generating valid Chemasim syntax with LLMs}

The core functionality of the Chemasim modelling agent lies in generating valid Chemasim syntax for the implementation of the process design.
 As discussed above, the underlying LLMs are generally not pre-trained on Chemasim syntax due to the very limited public exposure of the tool.
 Consequently, the key to enabling valid syntax generation lies in so-called in-context learning.

\paragraph{In-context learning for domain-specific languages}

We provide the LLM with relevant information as context in the form of technical documentation, commented examples and how-tos directly within the prompt - without any re-training or fine-tuning of the underlying model.
 Following this approach, the most recent generation of LLMs can reliably generate valid Chemasim syntax.
 In particular, models such as Claude Sonnet and Opus (starting from versions 4.5) as well as GPT (starting from version 5.2) are able to generate valid Chemasim syntax for a variety of unit operations and specification types mostly in one shot.
 Notably, older versions regularly produced substantial syntax errors, or even suggested pseudocode that did not follow the Chemasim syntax at all.
 In this context, we highlight that the most recent generation of LLMs inherently understands the concept of specifications, i.e., that setting a certain (originally computed) quantity to a fixed value requires to set another quantity (that needed a fixed value by default) free.
 We note that, in principle, entire flowsheet models comprising multiple interconnected units could be generated at once.
 However, as will be discussed below, a unit-by-unit implementation is strongly preferred in practice and indeed enforced via the Chemasim modelling agent's instructions to facilitate convergence of the simulation.

\paragraph{Provided context}

We provide the LLM with an extensive technical documentation of Chemasim, which is organized in a unit-based manner, i.e., describing the syntax for each unit operation and the different alternatives for specification in detail.
 This documentation further includes a description of Chemasim's functional language for defining more complex specifications (cf. the example in the previous section).
 As the technical documentation largely lacks comprehensive unit syntax examples with inline comments and preliminary explanations, we additionally provide how-tos for the different unit operations relevant to this work, covering their syntax and specification alternatives comparable to the illustrative presentation in the previous section.
 Lastly, we include commented full-flowsheet examples - specifically, Chemasim implementations of the processes presented by Luyben as mentioned above - to demonstrate the interconnection of multiple unit operations within a complete process model.
 The provided context files are available in the supplementary information.

\subsection{Agentic AI framework for autonomous Chemasim model development}

The basic AI framework comprising an LLM with context as described above is readily able to generate valid Chemasim syntax for many different unit operations and specification types.
 However, it cannot execute the resulting code and react to error messages or simulation results. This is achieved by integrating the LLM into an agentic AI framework with access to the Chemasim project.

\paragraph{Manipulating Chemasim input files}

As Chemasim models are defined in a textual manner, the agent can directly inspect, interpret, and manipulate the input files to implement the generated code.
 Thereby, the agent is readily able to, e.g., add new unit operations, modify existing ones, add additional specifications or deactivate existing ones.
 We again emphasize that this intuitive textual way of manipulating the flowsheet model is not common when using commercially available flowsheet simulators, where the graphical user interface is the primary way to build and modify flowsheet models.
 In that case, the agent would likely need to interact with an application programming interface (API), that would need to be provided and maintained by the software vendor and would likely be somewhat limited in terms of the possible interactions.
 In contrast, Chemasim's text-based approach allows for the most flexible interaction giving the agent the same possibilities as a human user would have.

\paragraph{Running simulations and analysing error messages}

An extension is developed to directly trigger runs of the Chemasim simulation engine from the text editor.
 The extension can also be used by the agent, allowing it to autonomously run simulations and read and interpret the console output.
 Thereby, the agent recognizes error messages that - particularly in the case of syntax errors - often include precise information on the location of the error.
 This allows the agent to directly react to these messages by correcting the syntax error and running the simulation again.

Chemasim has powerful built-in initialization routines for all unit models, which are used to generate initial guesses for the solver if a new unit is added to the flowsheet.
 Afterwards, the results of a previous simulation, a so-called profile, can be used as initialization for the solver.
 As a consequence, Chemasim offers - besides the standard "run" option that takes the last stored profile - a "re-run" option (German "Nachstart") for the simulation, which always first copies the results of the previous simulation to the initial profile.
 This option is on the one hand especially useful for preventing convergence issues when setting up a large process model in a unit-by-unit approach.
 On the other hand, it prevents the user from accidentally getting stuck in a non-converged solution as the overwriting needs to be explicitly triggered by the user.
 Knowing about these concepts and being able to use them appropriately is crucial for the design of a tailored Chemasim modelling agent that can effectively interact with the simulation software. 

\paragraph{Retrieving simulation results}

The Chemasim simulation engine writes the results of the simulation into an SQLite database, which could, in principle, be directly accessed by the agent to retrieve the results of the simulation.
 However, we also offer a Python package wrapping the database access and SQLite queries into more user-friendly functions, which can be used by the agent to retrieve the results of the simulation in a convenient way.
 Thereby, the agent can (i) retrieve the simulation status (converged, not converged, etc.) and react accordingly, and (ii) retrieve the values of arbitrary quantities of the model (flow rates, temperatures, compositions, etc.) and assess if the process behaves as expected.
 
\subsection{Chemasim modelling agent}

Having the ability to generate valid Chemasim syntax using context and interact with the simulation software is a crucial step towards an autonomous model development.
 Nevertheless, the LLM at the core of the agentic AI framework still lacks knowledge about the effective handling of the process simulation engine.
 To this end, we design a custom Chemasim modelling agent with specific instructions on how to interact with the Chemasim project.
 These include on the one hand general instructions on how to use the extension for running simulations and the Python functions for retrieving results properly as well as on where to find the relevant context files for generating valid Chemasim syntax.
 On the other hand, we also provide rather specific instructions on how to set up a process model in Chemasim as well as best-practices for ensuring a convergent simulation.

\paragraph{Unit-by-unit procedure}

The agent is instructed to implement the process model strictly in a unit-by-unit approach.
 In order to ensure the agent follows this instruction, the agent is forced to first generate a to-do list for the implementation of the process model, which is then executed in a sequential manner.
 In each step, the agent makes a single change to the input file, runs the simulation, and reads the results.
 Only after successful runs, a re-run is triggered to store the solution as initial guess for the solver before advancing to the next unit.
 In this context, a single change can be the addition of a new unit operation, the modification of an existing one, the addition of a specification and deactivation of an existing one.

\paragraph{Best-practices for distillation columns}

Distillation columns are a typical source of convergence issues in process simulations.
 When adding a new column, the agent is therefore instructed to strictly follow a two-stage specification procedure for each column.
 First, a column with a given number of stages is set up with a fixed bottom mass flow rate and a fixed reflux ratio, which corresponds to the most robust way for the solver to find a solution from scratch.
 After adding the column, the agent reads the simulation results and evaluates if the separation performance of the column is sufficient according to the process requirements.
 If requirements are not met, the agent is allowed to modify the specifications of the column, e.g., increasing the number of stages and/or reflux ratio to enhance the separation performance.
 Once the separation performance is sufficient, the agent is instructed to re-specify the column by replacing the specifications of bottom mass flow rate (and optionally reflux ratio) with the desired purities.
 This way of specifying the column is typically beneficial for the convergence of the simulation when closing recycle loops.
 As an alternative, the agent can also replace the specification of the bottom mass flow rate with a specification of the distillate/bottom-to-feed ratio, which can also be beneficial for convergence when closing recycle loops.

\paragraph{Best-practices for auxiliary materials and recycle streams}

Some process concepts include the recycling of large quantities of auxiliary materials.
 In that case, only small make-up flow rates are required to compensate the losses in the process.
 Here, we find it beneficial to first set up the process with a fixed flow rate of the anticipated auxiliary material recycle.
 When closing the recycle, the agent is instructed to make sure that the make-up flow is set free and computed to keep the total flow of the auxiliary material at a constant level and only compensate losses.

Regarding the general recycle structure, the agent shall always set up the open-loop process with the recycle being treated as external feed.
 In this context, it is, obviously, particularly advantageous to have estimates that are as close as possible to the final solution for the flow rates and compositions of the recycle stream.
 Ensuring high-quality initial guesses is, in the remainder of this work, exclusively done by the process development agent.

In case of convergence issues when closing the recycle, the agent is instructed to first check adherence to the best-practices for distillation columns and auxiliary material recycle as described above.

\subsection{Process development agent}

The process development agent generates the unit-by-unit process documentation that is processed by the Chemasim modelling agent for obtaining the to-do list for implementation.
 Moreover, the process development agent provides the necessary information for the implementation of each unit operation as well as estimates for relevant flow rates and compositions.
 For distillation columns, the agent is also instructed to provide preliminary estimates for the number of stages and the reflux ratio.
 For these purposes, the agent has access to tools for thermodynamic calculations to analyze the system and to perform mass balance computations in free Python coding to estimate flow rates and compositions.
 Except for a few guiding questions for reasoning the design, the process development agent is not given any specific instructions on how to solve the process synthesis problem.
 Neither is it given any further process engineering knowledge beyond its inherent understanding of the underlying phenomena.

\paragraph{Tools for thermodynamic calculations}

The process development agent has access to a set of tools for performing thermodynamic calculations that are essential for the analysis of the given system and thus for the design of the process flowsheet.
 In particular, the agent can compute azeotropes for multi-component systems at a specified pressure, which also provides information on the type of the azeotrope (minimum or maximum boiling, homogeneous or heterogeneous), its composition, and the corresponding boiling temperature.
 Furthermore, the agent can compute binary and ternary miscibility gaps at a given temperature. In the binary case, this includes the compositions of the coexisting phases, whereas in the ternary case, a set of tie lines is returned as well as the critical point of the system. 
 In the case of reactive processes, the agent can additionally parse the reaction network file from the project to extract the stoichiometry, reactants and products of each possible reaction.

\paragraph{Guiding questions for reasoning the design}

As stated above, the process development agent is not given any specific instructions on how to solve the process synthesis problem and mostly relies on its inherent knowledge.
 We only provide a few guiding questions to ensure that the agent follows a structured reasoning approach and critically challenges its design decisions.
 These questions cover, among others, the following aspects:
 \begin{itemize}
  \item Recycle structure: Which recycles are needed to increase conversion or recovery? If adding auxiliary materials, can they be recycled to reduce the make-up flow? Can throughput be reduced? Is a purge stream foreseen to prevent the accumulation of impurities?
  \item Distillation sequences: Which product can be withdrawn in a column? Do azeotropes prohibit sharp splits? What is the effect of an adjusted pressure? Can recycling of pure components or adding auxiliary material shift the feed in a beneficial manner? Is the separation rather difficult, requiring many stages and a high reflux ratio?
  \item Miscibility gaps: Are there any miscibility gaps to be exploited? Would a multi-stage separation be beneficial?
  \item Overall process design: Is the sequence of units reasonable? Are there units that are not strictly required? Can multiple units be combined into one unit? 
 \end{itemize}
 Having suggested a process design based on the guiding questions, the agent is then asked to critically evaluate its design, suggest possible improvements, and discuss potential alternatives. These can, among others, target reduced losses, reduced throughput, reduced number of units, etc.

\paragraph{Mass balance computations}

The agent has the ability to freely code in Python, which it is instructed to use for performing mass balance computations to estimate flow rates and compositions of the different streams in the process.
 Here, the agent, without further instructions, relies on simplifying assumptions, such as sharp splits in distillation columns, to perform the computations that mostly lead to solving a system of linear equations.
 If necessary, the agent can also perform more complex computations, requiring iterative approaches.

\subsection{Implementation details}

We use Visual Studio Code (VS Code)~\citep{vscode} as the platform for the implementation of the multi-agent system with a custom extension realizing the interface to the Chemasim project and the simulation engine.
 Using VS Code further allows for a straightforward application of GitHub Copilot~\citep{githubCopilot} as basis for the agentic AI framework.
 All results presented in this work are obtained using Claude Opus 4.6~\citep{anthropicClaude} as the underlying LLM for both agents, which is considered the most proficient model for that task at the time of writing.
 As indicated in Figure~\ref{fig:mas}, the user can intervene in the process by requesting changes to the process design.
 Furthermore, there is a technical possibility for feedback from the Chemasim modelling agent to the process development agent, which is, however, barely used in this work.
 System prompts for both agents can be found in the supplementary information.

\section{Results and discussion}

We test the developed multi-agent system on multiple process development case studies that require very common process engineering knowledge and techniques.
 We emphasize that in all cases, we mask the components as A, B, C, etc. (cf. Table~\ref{tab:case_study_components}) to ensure that the process development agent cannot use specific knowledge about the components and the respective processes.
 Moreover, we initially do not give any specific instructions on how to solve the process synthesis problem and solely rely on the inherent knowledge of the agent and its capabilities to reason about the guiding questions.
 We also highlight that the process development agent is not given the context for Chemasim to prevent it from transferring any knowledge from the examples provided to the Chemasim modelling agent to the process design.
 The resulting Chemasim models as well as the simulation results in form of stream tables can be found in the supplementary information.

\begin{table}[htbp]
\centering
\caption{Component mapping for the masked case studies in the Results and discussion section.}
\label{tab:case_study_components}
\begin{tabular}{llllll}
\toprule
Case & A & B & C & D & E \\
\midrule
1 & Ethylene & Benzene & Ethyl & Diethyl & -- \\
& &  & benzene & benzene & \\
2a & n-Propanol & Benzene & -- & -- & -- \\
2b & Acetone & Chloroform & -- & -- & -- \\
3 & Water & 1,4-dioxane & Benzene & Dimethyl & -- \\
& & & & carbonate & \\
4 & Water & Pyridine & Dimethyl & n-Octanol & Toluene \\
& & & carbonate & & \\
5 & n-Propanol & Benzene & Toluene & -- & -- \\
\bottomrule
\end{tabular}
\end{table}

\subsection{Case study 1: Reaction-separation process}

This case study is inspired by the ethyl benzene process presented by \cite{Luyben2011b}. That is, we consider the formation of ethyl benzene (component C) from benzene (component B) and ethylene (component A).
 We further consider the formation of the by-product diethyl benzene (component D) as well as a transalkylation reaction to decompose diethyl benzene in the presence of benzene back to ethyl benzene.

We target a production of 1000~kg/h of C. The process development agent receives the following prompt:

\begin{agentprompt}
Suggest a process to produce the given stream having A (gas at 25°C and 1~atm) and B (liquid at 25°C and 1~atm) as feed components.
Reactions take place in the liquid phase at temperature of 180°C and pressure of 20~bar.
B is in excess (molar ratio of 2 to 1).
Overall conversion of A is 100\% with selectivity of 90\% towards C.
Side product D can be decomposed to form C at higher temperature (240°C) and 20~bar at high excess of B (molar ratio of 5 to 1).
Conversion in this step is 80\%.
This requires very pure D.
Further, quick cooling to below 150°C afterwards is required.
For separation in distillation columns, bottom temperature should not exceed 120°C.
\end{agentprompt}

\paragraph{Analysis of the system}

Table~\ref{tab:reactions_case1} summarizes the given reaction network as read by the agent from the project files. We further add the extracted relevant information from the prompt.

\begin{table}[htbp]
	\centering
	\caption{Reaction network and operating conditions for case study 1. All reactions occur in the liquid phase.}
	\label{tab:reactions_case1}
	\begin{tabular}{llccl}
		\hline
		Reaction & Stoichiometry & $T$ & $p$ & Conditions \\
		 & & [°C] & [bar] & \\
		\hline
		Main & A + B $\rightarrow$ C & 180 & 20 & 90\% conversion w.r.t. A \\
		 & & & & 2:1 molar B:A ratio \\[4pt]
		Side & A + C $\rightarrow$ D & 180 & 20 & 10\% conversion w.r.t. A \\
		 & & & & \\[4pt]
		Transalkylation & D + B $\rightarrow$ 2\,C & 240 & 20 & 80\% conversion w.r.t. D \\
		 & & & & 5:1 molar B:D ratio, pure D required \\
		\hline
	\end{tabular}
\end{table}

As a second step, the agent computes the boiling sequence of the pure components at reasonable pressures (cf. Table~\ref{tab:boiling_case1}), i.e., those that do not exceed the maximum bottom temperature of 120°C.
 From this, the agent can already start to reason about possible separation sequences which are straightforward due to the absence of azeotropes.

\begin{table}[htbp]
	\centering
	\caption{Boiling points of pure components A--D at selected pressures for case study 1.}
	\label{tab:boiling_case1}
	\begin{tabular}{ccccc}
		\hline
		$p$ & $T_\mathrm{b}$(A) & $T_\mathrm{b}$(B) & $T_\mathrm{b}$(C) & $T_\mathrm{b}$(D) \\
		bar & °C & °C & °C & °C \\
		\hline
		0.12 & $-133.0$ & 23.9  & 71.4  & 112.9 \\
		0.15 & $-130.5$ & 28.7  & 77.0  & 119.1 \\
		0.25 & $-124.3$ & 40.7  & 90.8  & 134.1 \\
		0.50 & $-114.9$ & 58.8  & 111.7 & 157.0 \\
		1.00 & $-104.0$ & 79.7  & 135.7 & 183.3 \\
		\hline
	\end{tabular}
\end{table}

\paragraph{Process design}

\begin{figure}[htbp]
	\centering
	\includegraphics[width=\textwidth]{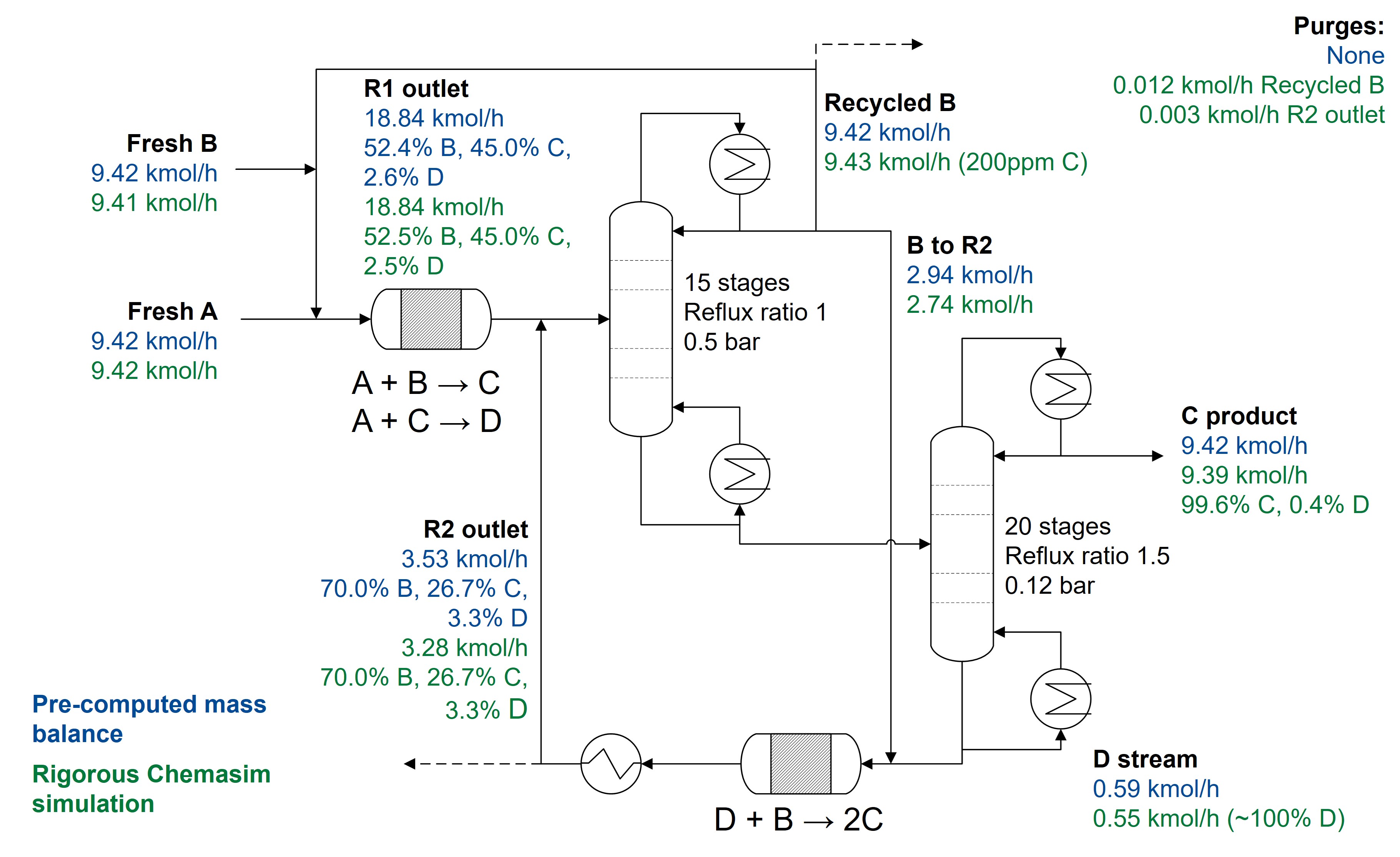}
	\caption{Final process flow diagram for case study 1 - reaction-separation process. Blue numbers show the molar balance as pre-computed by the process development agent. 
	Green numbers show the results from a rigorous simulation of the flowsheet built by the Chemasim modelling agent. Dashed lines indicate purge streams planned by the Chemasim modelling agent to prevent the accumulation of impurities in the recycle.
	If no composition is given, the process development agent assumes a pure component stream.}
	\label{fig:pfd_case1}
\end{figure}

Based on the analysis of the system, the agent designs the process flowsheet as illustrated in Figure~\ref{fig:pfd_case1}.
 The process includes a first reactor for the main and side reaction at given conditions (180°C, 20~bar, 2:1 molar B:A ratio).
 Assuming full conversion of A in total, no recycle of unconverted A that might involve the treatment of a gas phase is required.
 The reaction mixture is fed to a first distillation column operating at 0.5~bar to withdraw pure B as top product to be recycled to the reactor.
 B-free bottom product is fed to a second distillation column operating at lower pressure of 0.12~bar to separate C as top product from D as bottom product adhering to the maximum bottom temperature constraint of 120°C.
 The pure D is then fed to a second reactor for the transalkylation reaction at given conditions (240°C, 20~bar, 5:1 molar B:D ratio) to produce more C.
 Feed B for this reactor is withdrawn as side stream from the B recycle from the top of the first column.
 The effluent of the second reactor is cooled to below 150°C and enters the first distillation column together with the effluent of the first reactor to separate B for recycling. 

As can be seen from the blue figures in the process flow diagram (Figure~\ref{fig:pfd_case1}), the process development agent is able to perform mass balance computations to estimate flow rates and compositions of the different streams in the process.
 Even though the agent's estimates are based on simplifying assumptions, such as sharp splits in distillation columns, they provide very valuable initial guesses for the Chemasim modelling agent to set up the process model and ensure convergence of the simulation, which will be discussed in more detail below.

\paragraph{Rigorous simulation}

Based on the documentation provided by the process development agent, the Chemasim modelling agent is able to implement the process model following its instructions as described above.
 Simulation results are given in the green figures in the process flow diagram (Figure~\ref{fig:pfd_case1}) and show very good agreement with the estimates of the process development agent.
 As the only additional instruction, we ask the Chemasim modelling agent to adhere strictly to the reaction conditions. This leads to the agent freeing the fresh B inlet and the flow rate of the B recycle stream to the second reactor.
 In turn, the agent adds specifications to keep the desired molar ratios of B to A and D in the two reactors and compensate for B losses in the product or via purge stream. Without these specifications, small deviations in the fresh B flow rate could potentially blow up the recycle flow.

As further specifications, the agent uses a B impurity of 500~ppm in the feed to the second column to prevent contamination of the C product stream, for which a specification of 99.5\% (mass, approx. 99.6\% molar) purity is selected.
 Note that as instructed, these purity specifications are added after the initial setup of the column with specifications of reflux ratio and bottom mass flow rate (replaced).
 Thereby, the agent ensures a convergent simulation when closing the recycle loop.

\subsection{Case study 2: Pressure-swing distillation}

The second case study focuses on the design of separation sequences for binary azeotropic systems, where the azeotropic composition exhibits a significant pressure-dependence.
 Here, we consider two thermodynamic systems with different types of azeotropes. First, we consider a binary mixture of n-propanol (component A) and benzene (component B) forming a minimum boiling azeotrope, which has been studied as a subsystem of a quinary system \citep{Liu2004,Sasi2020}.
 Second, we consider a binary mixture of acetone (component A) and chloroform (component B), which typically serves as example for a maximum boiling azeotrope.
 Here, pressure-swing distillation has been proposed as a separation strategy, however, extractive distillation with an entrainer is commonly preferred \citep{Luyben2013}.
 Nevertheless, we use this system as a test case for the design of pressure-swing distillation sequences to demonstrate the capabilities of the multi-agent system for coping with different thermodynamic behavior.
 Note that, with the current limitation to considering pre-defined components, the agent is in fact in this case study not able to design, e.g., an extractive distillation sequence with an entrainer.

In both cases, we apply bounds on temperature for testing purposes. The prompts read:

\begin{agentprompt}
Suggest a process to separate this stream. Do not go to vacuum and stay below 130°C in distillation columns.
\end{agentprompt}

for case 2a (250~kg/h n-propanol (A) and 250~kg/h benzene (B)) and

\begin{agentprompt}
Suggest a process to separate this stream. Do not go to strong vacuum below 0.5 bar and stay below 150°C in distillation columns.
\end{agentprompt}

for case 2b (500~kg/h acetone (A) and 500~kg/h chloroform (B)).

\paragraph{Analysis of the system}

\begin{figure}[htbp]
	\centering
	\begin{subfigure}[b]{0.48\textwidth}
		\centering
		\includegraphics[width=\textwidth]{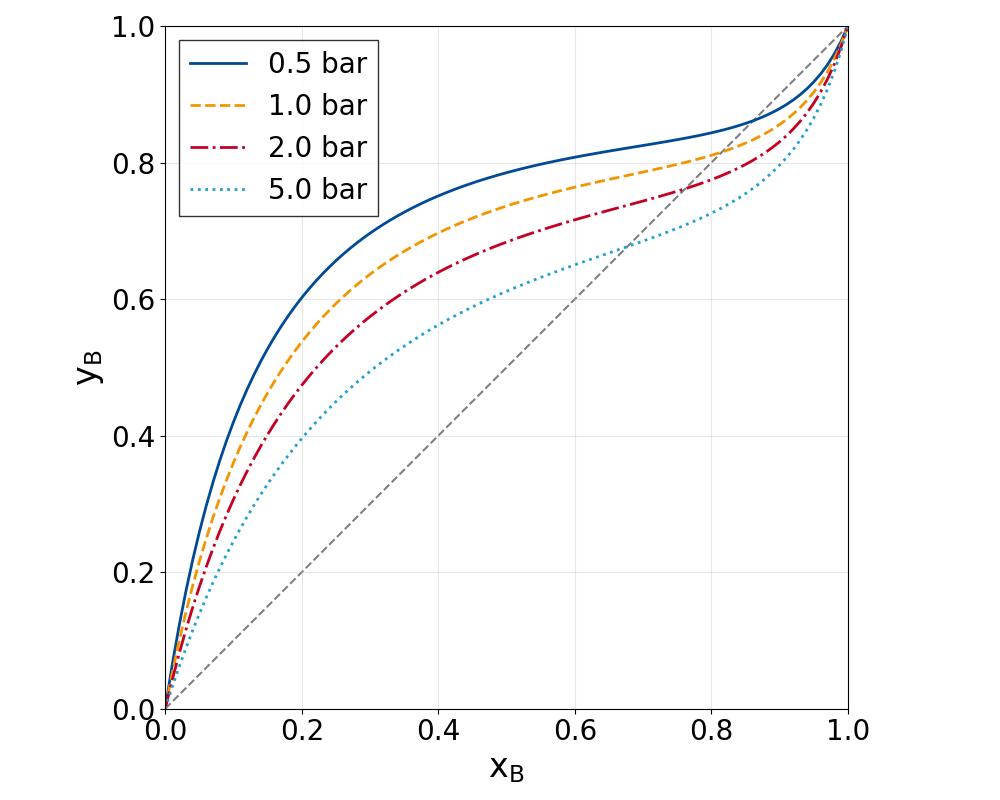}
		\caption{Binary A-B system with B as light boiler forming a minimum azeotrope (case study 2a).}
		\label{fig:psd_left}
	\end{subfigure}
	\hfill
	\begin{subfigure}[b]{0.48\textwidth}
		\centering
		\includegraphics[width=\textwidth]{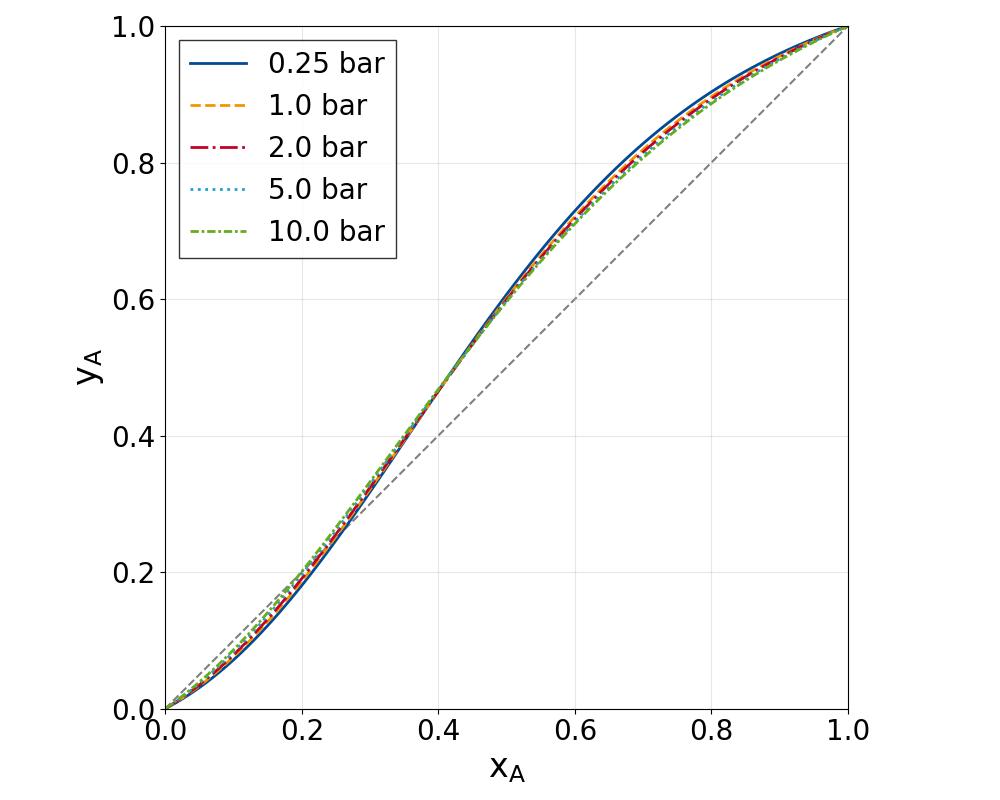}
		\caption{Binary A-B system with A as light boiler forming a maximum azeotrope (case study 2b).}
		\label{fig:psd_right}
	\end{subfigure}
	\caption{Pressure-dependent xy-diagrams for case study 2 at selected pressures. Colors and line-styles indicate different pressures as given in the legend.}
	\label{fig:psd}
\end{figure}

The main insight from the thermodynamic analysis of the systems lies in the pressure-dependence of the azeotropic composition as illustrated in the pressure-dependent xy-diagrams in Figure~\ref{fig:psd}.
 The corresponding boiling points of the pure components and the azeotrope at the selected pressures are given in Table~\ref{tab:psd_boiling}.
 As can be seen, system 2a exhibits a much more significant pressure-dependence of the azeotrope than system 2b. Moreover, the separability of the components by distillation is much better with a larger difference in the boiling points of the pure components in system 2a and a wider two-phase region (i.e., a larger distance to the parity line in the xy-diagram).
 
\begin{table}[htbp]
\centering
\caption{Pure-component boiling points and azeotrope temperatures for case study 2 at selected pressures. MI and MA refer to minimum and maximum azeotrope, respectively.}
\label{tab:psd_boiling}
\begin{tabular}{lcccc}
\toprule
Case study & $p$ [bar] & $T_\mathrm{b}$ (A) [°C] & $T_\mathrm{b}$ (B) [°C] & $T_\mathrm{azeo}$ [°C] \\
\midrule
2a (MI) & 0.5  & 79.82 & 58.80 & 56.79 \\
	& 1.0  & 97.07 & 79.66 & 76.13 \\
	& 2.0  & 116.65 & 103.91 & 98.05 \\
	& 5.0  & 147.08 & 142.70 & 131.90 \\
\midrule
2b (MA) & 0.25 & 20.26 & 23.86 & 27.03 \\
	& 1.0  & 55.68 & 60.57 & 63.97 \\
	& 2.0  & 77.46 & 83.39 & 86.83 \\
	& 5.0  & 111.86 & 119.74 & 123.11 \\
	& 10.0 & 143.22 & 153.12 & 156.31 \\
\bottomrule
\end{tabular}
\end{table}

Furthermore, the agent analyses which column pressures adhere to the temperature constraints as stated in the prompt.
 Here, for system 2a, staying below 130°C allows for column pressures of up to approx. 3~bar, which would lead to a boiling temperature of 129.4°C for the pure component A as maximum boiling temperature of the binary system.
 For system 2b, pressures of lower than 9~bar are allowed, which would cause an azeotropic boiling temperature of 150.9°C as maximum.

\paragraph{Process design}

\begin{figure}[htbp]
	\centering
	\includegraphics[width=\textwidth]{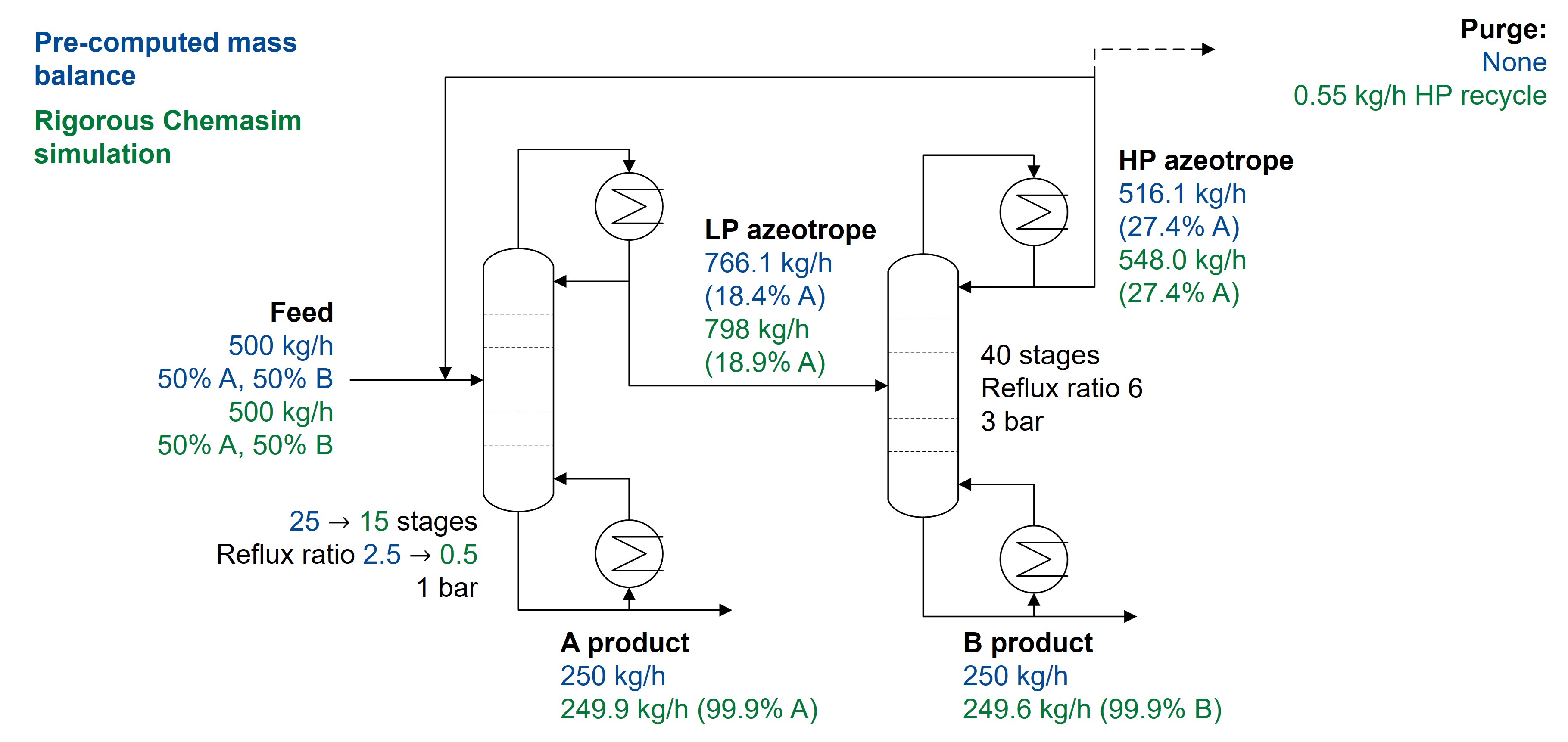}
	\caption{Final process flow diagram for case study 2a - binary pressure-swing distillation for a minimum azeotrope. Blue numbers show the mass balance as pre-computed by the process development agent. 
		Green numbers show the results from a rigorous simulation of the flowsheet built by the Chemasim modelling agent. Dashed lines indicate purge streams planned by the Chemasim modelling agent to prevent the accumulation of impurities in the recycle.
		If no composition is given, the process development agent assumes a pure component stream.}
	\label{fig:pfd_case2}
\end{figure}

Besides the type of azeotrope, the relative location of the feed composition to the azeotropic composition is decisive for the design of the separation sequence.
 Presenting the 50/50 feed mixture of system 2a as example (cf. Figure~\ref{fig:pfd_case2}), the feed is located on the A-rich side of the azeotrope at all pressures.
 Consequently, pure A can be withdrawn as bottom product from the first column, whereas the minimum boiling azeotrope is withdrawn as top product.
 As the azeotropic composition shifts to the A-rich side at higher pressures, the first column operates at the lower pressure.
 In that case, the azeotrope withdrawn as top product and fed to the second column is located on the B-rich side of the azeotrope at the elevated pressure.
 Consequently, pure B can be withdrawn as bottom product from the second column, whereas the minimum boiling azeotrope is withdrawn as top product and recycled to the first column.

Notably, even though the agent would be allowed by the prompt to go to pressures as low as 0.5~bar to maximize the shift in azeotropic composition and further enhance the separation as Fig.~\ref{fig:psd_left} suggests, it selects a pressure of 1.~bar for the first column.
 Thereby, the agent intentionally omits vacuum operation, which appears to be a reasonable design decision considering the in either case relatively good separability of the components that is confirmed by low reflux and number of stages in the rigorous simulation.
 Note however that the final selection of column pressure is subject to a certain degree of variability which particularly holds if relaxing the temperature constraint in the prompt.

We herein omit a detailed discussion of the process design for system 2b, which is overall similar to system 2a, but with the differences that (i) the first column operates at the higher pressure and the second column at the lower pressure, and (ii) pure components are withdrawn as top products and the azeotropes are withdrawn as bottom products.
Moreover, the agent accounts for the much narrower two-phase region in system 2b by selecting higher numbers of stages and reflux ratios. Likewise, the smaller shift in azeotropic composition leads to a much higher recycle flow from the second column to the first column foreseeing a recycle-to-feed ratio of approx. 2.

\paragraph{Rigorous simulation}

As in the previous case study, the process models implemented by the Chemasim modelling agent based on the process design documentation provided by the process development agent lead to simulation results that are in very good agreement with the estimates of the process development agent.
 Here, we emphasize that under the given assumptions of the mass balance computations, the process development agent estimates the minimum recycled amount of high-pressure azeotrope.
 Achieving a less sharp split in a rigorous distillation column model thus causes the higher recycle flow.

As can be further seen, the Chemasim modelling agent reduces the number of stages and the reflux ratio in the first distillation column to avoid over-separation (however, mostly from a convergence perspective and not aiming at an economic optimization).
 Along these lines, the much higher reflux ratio and number of stages in the second column (that follows from the narrower two-phase region at higher B concentrations) also appear somewhat excessive as the azeotropic composition is indeed reached at the top.
 However, the agent does not reduce these specifications as it appears to be acceptable for convergence of the simulation and as the agent is not given any specific instructions on optimizing the design.

Notably, in case of system 2b, a much stronger correction towards higher numbers of stages and reflux ratios is observed to achieve the desired separation with, e.g., the second column comprising 120 stages and a reflux ratio of 40.
 This worse separability of the components is also reflected in a more significant deviation from the estimated minimum recycle flow (assuming maximum splits in the column) increasing the recycle-to-feed ratio to >3.

\subsection{Case study 3: Heteroazeotropic distillation with entrainer selection}

In the third case study, we target the design of a separation sequence for a binary mixture of water (component A) and 1,4-dioxane (component B).
 We first explicitly consider a heteroazeotropic distillation sequence with an entrainer, where we give two options (components C and D) to the agent.
 Here, component C is benzene, which has been proposed as entrainer for this system in the literature \citep{Kruber2022} that offers a very large miscibility gap.
 Component D is dimethyl carbonate and represents an alternative entrainer.

The original prompt for the process development agent targets separating a mixture of 1000~kg/h A and 1000~kg/h B:

\begin{agentprompt}
Suggest a process to separate this stream. Do not change from ambient pressure.
\end{agentprompt}

As will be shown below, this prompt leads to the design of a heteroazeotropic distillation sequence using C as entrainer with one distillation column and a decanter.
 In order to test the capabilities of the agent system, we also drive it to design a process in line with the design by \cite{Kruber2022} using two distillation columns.
 This is done using the following prompt:

\begin{agentprompt}
Reconsider the process design from the previous version. Explicitly prefer variants that lower operating costs by reducing throughput and thus energy demand, even if this requires additional equipment.
\end{agentprompt}

Last but not least, we also let the agent design the separation sequence more freely, omitting the specific instruction to stay at ambient pressure, and thus allowing for pressure-swing distillation sequences as was in fact conceptually studied by \cite{Wu2014}.
 To this end, a third prompt is given to the agent:

\begin{agentprompt}
Reconsider your process designs. If you were allowed to adjust the pressure, what would you do?
\end{agentprompt}

\paragraph{Analysis of the system}

\begin{figure}[htbp]
	\centering
	\begin{subfigure}[b]{0.48\textwidth}
		\centering
		\includegraphics[width=\textwidth]{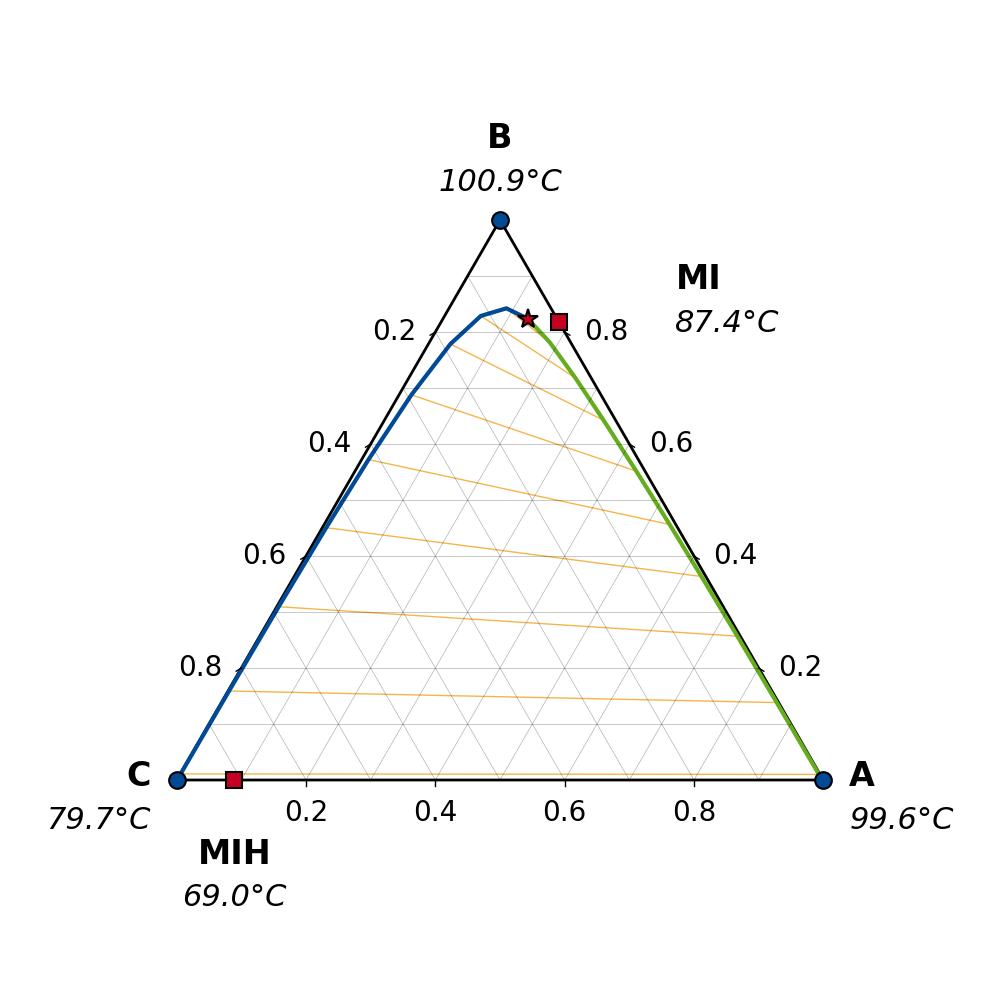}
		\caption{Ternary system for entrainer C.}
		\label{fig:tern_left}
	\end{subfigure}
	\hfill
	\begin{subfigure}[b]{0.48\textwidth}
		\centering
		\includegraphics[width=\textwidth]{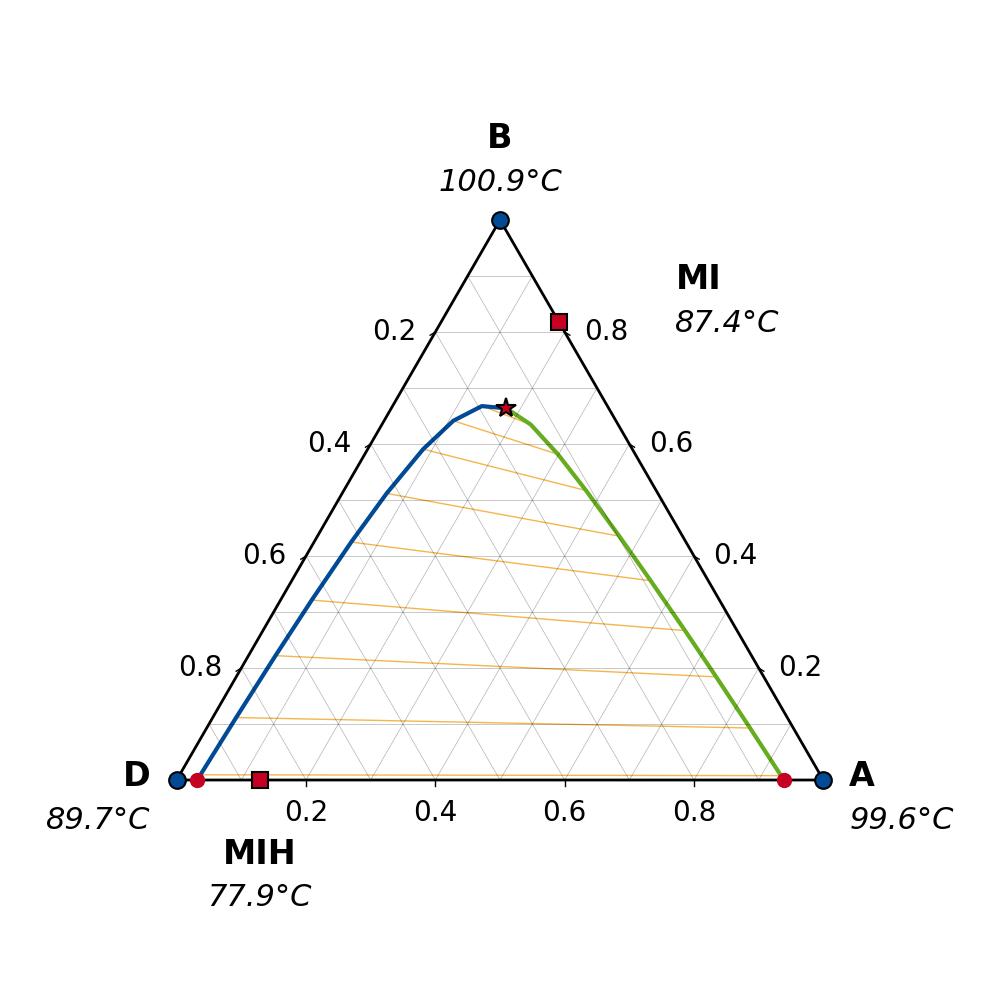}
		\caption{Ternary system for entrainer D.}
		\label{fig:tern_right}
	\end{subfigure}
	\caption{Ternary diagram for the heteroazeotropic distillation systems for separating A from B in case study 3. Diagrams include the azeotropes and (ternary) miscibility gaps. Squares mark azeotropes, with MI denoting homogeneous minimum azeotropes and MIH heterogeneous minimum azeotropes. Thin lines inside the miscibility gap show tie lines connecting liquid phases in equilibrium. The critical point of the liquid-liquid equilibrium is marked with a star.}
	\label{fig:tern_diox}
\end{figure}

In order to design the separation sequence, the agent first analyzes the ternary systems of A-B-C and A-B-D, which include the azeotropes and miscibility gaps as illustrated in the ternary diagrams in Figure~\ref{fig:tern_diox}.
 As can be seen, the challenge lies in overcoming the minimum boiling azeotrope of the binary A-B system at ambient pressure, which is located at the B-rich side of the feed composition.
 For the A-B-C system, the agent identifies a large miscibility gap with an A-C mixture splitting into very pure A and C phases.
 For the A-B-D system, the miscibility gap is much smaller, so that resulting phases from an A-D mixture are less pure and might require further processing.

Regarding the location of the A-entrainer azeotrope, we find rather similar compositions for the minimum heteroazeotropes in the A-C and A-D system.
 Consequently, the topology of the two ternary diagrams is similar with respect to the distillation regions and, hence, the entrainer amount required to shift the feed composition into the other region will be comparable.

\paragraph{Process design}

\begin{figure}[htbp]
	\centering
	\includegraphics[width=\textwidth]{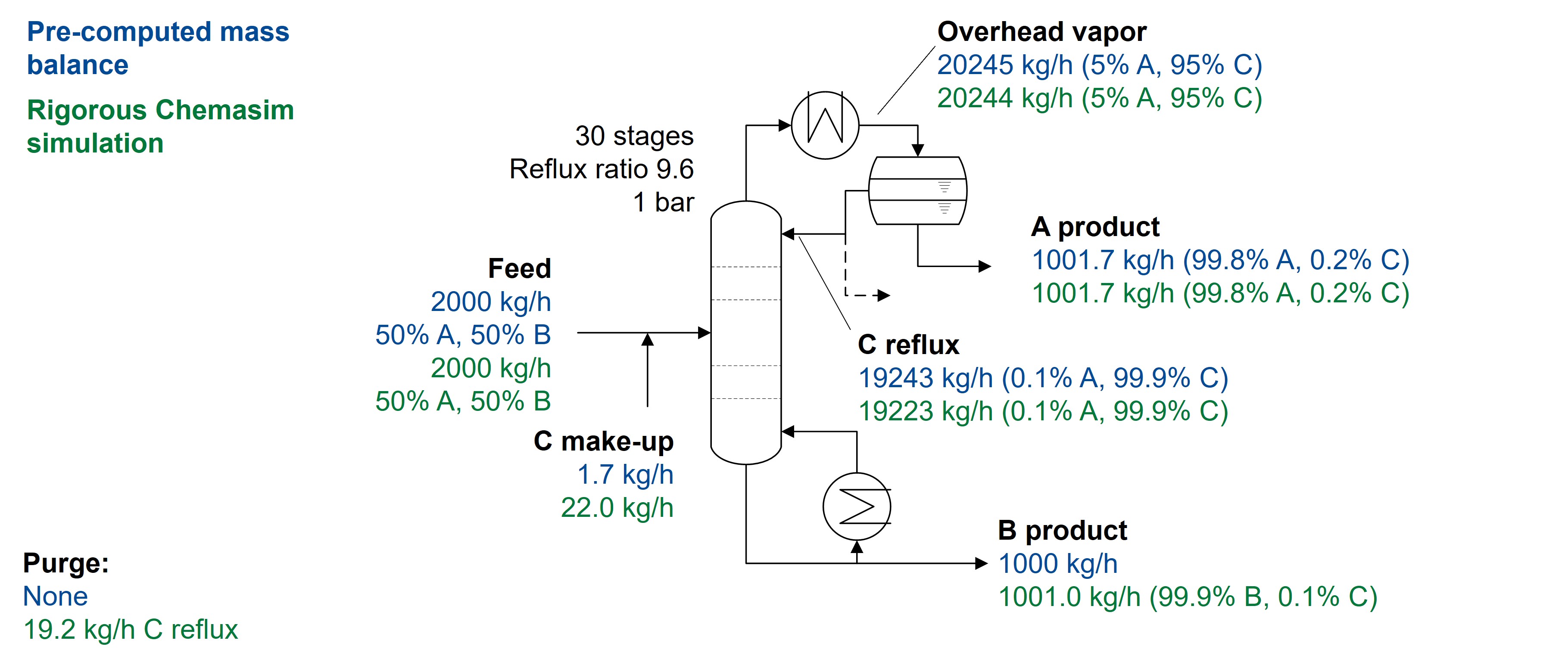}
	\caption{Process flow diagram for case study 3 as originally proposed by the agent system using a classic heteroazeotropic distillation approach with one distillation column and a decanter.
		Blue numbers show the mass balance as pre-computed by the process development agent. Green numbers show the results from a rigorous simulation of the flowsheet built by the Chemasim modelling agent. Dashed lines indicate purge streams planned by the Chemasim modelling agent to prevent the accumulation of impurities in the recycle.
		If no composition is given, the process development agent assumes a pure component stream.}
	\label{fig:pfd_case3a}
\end{figure}

We first of all again emphasize that components are masked as A, B, C, D, so that the agent cannot use specific knowledge about the components and the respective processes.
 Consequently, the agent bases the process design solely on the analysis of the thermodynamic behavior and cannot consider hazard potentials or the like.
 From that point of view, the entrainer C is clearly superior to D due to the much larger miscibility gap, which allows for a very efficient separation with only one distillation column and a decanter as depicted in Figure~\ref{fig:pfd_case3a}.

In that case, the A-B feed enters the distillation column at an intermediate stage, while almost pure C coming from the decanter is fed to the top as entrainer.
 Due to the unfavorable location of the binary A-C azeotrope close to the C vertex, a very large amount of entrainer is required to shift the feed composition in a way that allows for the withdrawal of pure B as bottom product.
 Here, we explicitly enforce an additional safety margin in the amount of C as otherwise the agent would target a sharp separation between pure B at the bottom and exactly the binary A-C azeotrope at the top, which would be numerically very challenging to achieve in a rigorous distillation column model.
 The vapor top product of the column is condensed and fed to a decanter, where it splits into a C-rich phase and an A-rich phase. The C-rich phase is recycled to the top of the column, whereas the A-rich phase is withdrawn as product in high purity.

\begin{figure}[htbp]
	\centering
	\includegraphics[width=\textwidth]{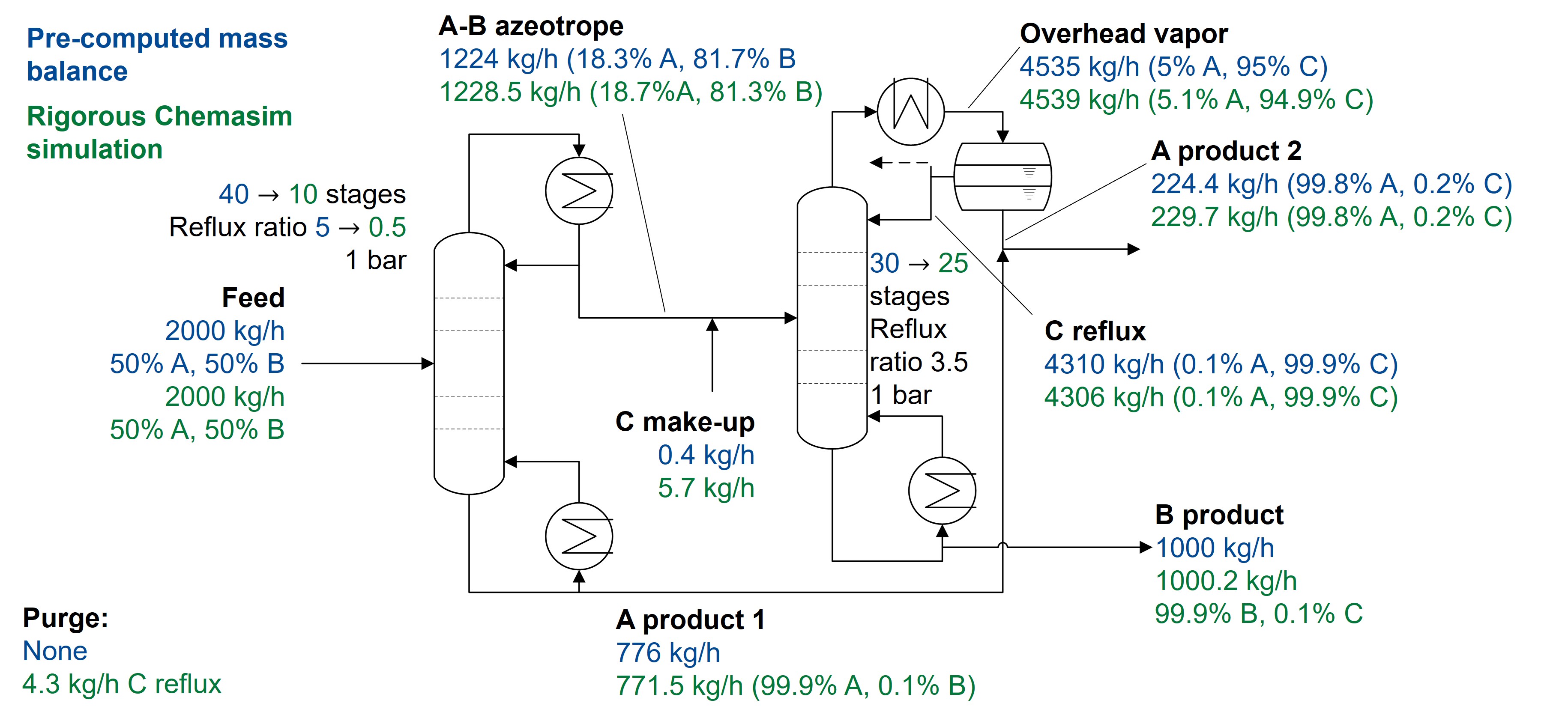}
	\caption{Process flow diagram for case study 3 enforcing a modified heteroazeotropic distillation approach with a pre-separation step. Blue numbers show the mass balance as pre-computed by the process development agent. Green numbers show the results from a rigorous simulation of the flowsheet built by the Chemasim modelling agent. Dashed lines indicate purge streams planned by the Chemasim modelling agent to prevent the accumulation of impurities in the recycle. If no composition is given, the process development agent assumes a pure component stream.}
	\label{fig:pfd_case3b}
\end{figure}

When requesting alternatives that omit the large C recycle, the agent adjusts the process design to a modified heteroazeotropic distillation approach with a pre-separation step as illustrated in Figure~\ref{fig:pfd_case3b}.
 Here, the feed is first fed to a distillation column, where pure A is withdrawn as bottom product and the A-B azeotrope leaves the column as top product.
 The A-B azeotrope is then fed to a second heteroazeotropic distillation column similar to the one above, which needs a substantially lower amount of entrainer due to the more favorable feed composition and overall lower flow.

If finally letting the agent freely adjust the pressure, it strongly favors a pressure-swing distillation sequence which would follow the process flow diagram depicted in Figure~\ref{fig:pfd_case2}.

\paragraph{Rigorous simulation}

Process designs for the third case study can also be robustly implemented by the Chemasim modelling agent and lead to simulation results that are in good agreement with the estimates of the process development agent.
 Here, the most noticeable deviations appear in the amount of C make-up that is foreseen to compensate for losses in the product stream (only the C content in the A-rich phase from the decanter in the simplified mass balance) and - more importantly, the loss via the purge stream in the rigorous simulation, which is required to prevent the accumulation of (organic) impurities.

Besides that, the figures for the pre-separation column in the modified heteroazeotropic distillation approach already indicate that this variant will likely be superior.
 In fact, against the original estimates, only very few stages and a very low reflux ratio are required in the first column.
 The second column then also benefits from a more favorable feed composition and lower entrainer reflux. In sum, one could thereby expect a reduction in energy demand of more than 60\% compared to the original design.
 At this point, we forego a more detailed analysis of the economic performance of the different designs - in particular, also regarding the comparison to the pressure-swing distillation sequence - as this is not the main focus of this work.

\subsection{Current limitations and directions for future work}

The presented case studies demonstrate the capabilities of the proposed multi-agent system for reasonable process design based on the analysis of the system and subsequent implementation of the process model in a rigorous process simulation environment.
 Nevertheless, the case studies also reveal some limitations of the current state of autonomous process development using agentic AI, which we discuss in the following and provide directions for future work.

\paragraph{Thermodynamic knowledge and tool usage}

The complexity level of the case studies is indeed very typical for the relevant literature with a focus on systems comprising two to four/five components and a few unit operations.
 Likewise, the considered thermodynamic behavior shows a certain degree of complexity, e.g., with the presence of azeotropes and miscibility gaps, but is still rather common.
 With increasing complexity of the system, we find that the agentic AI system is currently not able to fully cope with the analysis of the system and the design of appropriate processes.

\begin{figure}[htbp]
	\centering
	\begin{subfigure}[b]{0.32\textwidth}
		\centering
		\includegraphics[width=\textwidth]{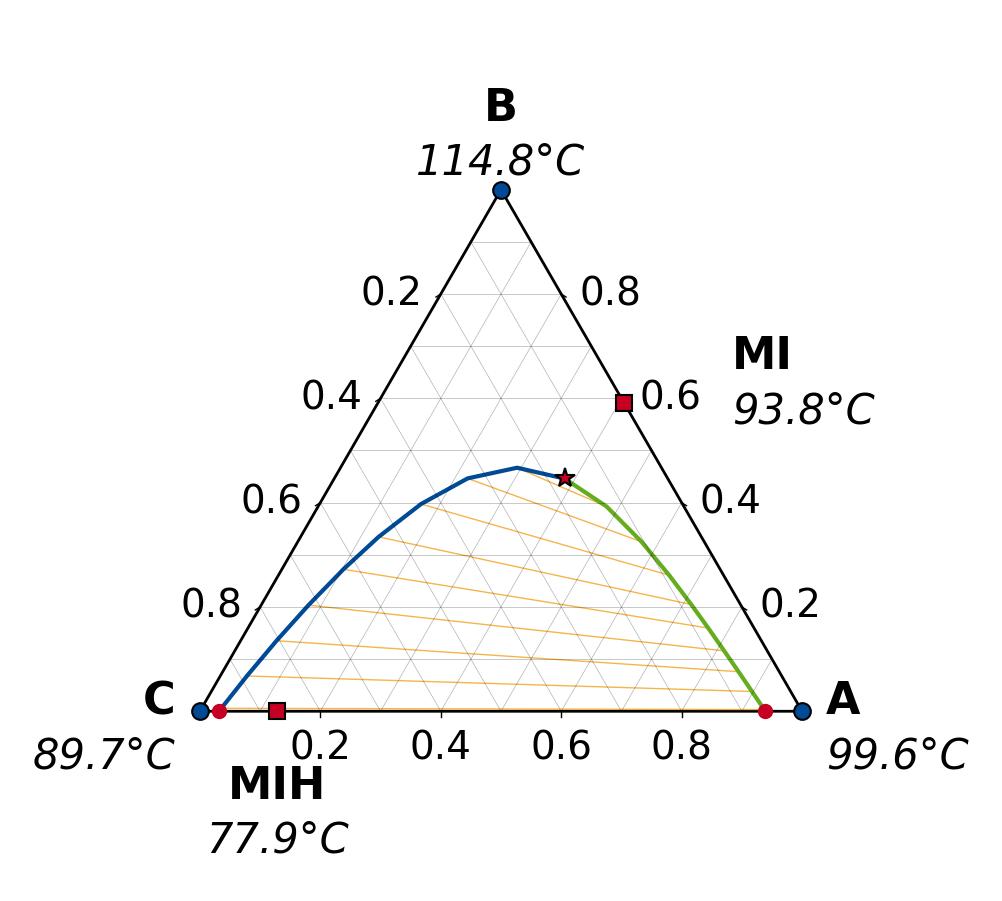}
		\caption{Entrainer candidate C with partial A-C miscibility requiring additional purification steps.}
		\label{fig:tern_pyr_dmc}
	\end{subfigure}
	\hfill
	\begin{subfigure}[b]{0.32\textwidth}
		\centering
		\includegraphics[width=\textwidth]{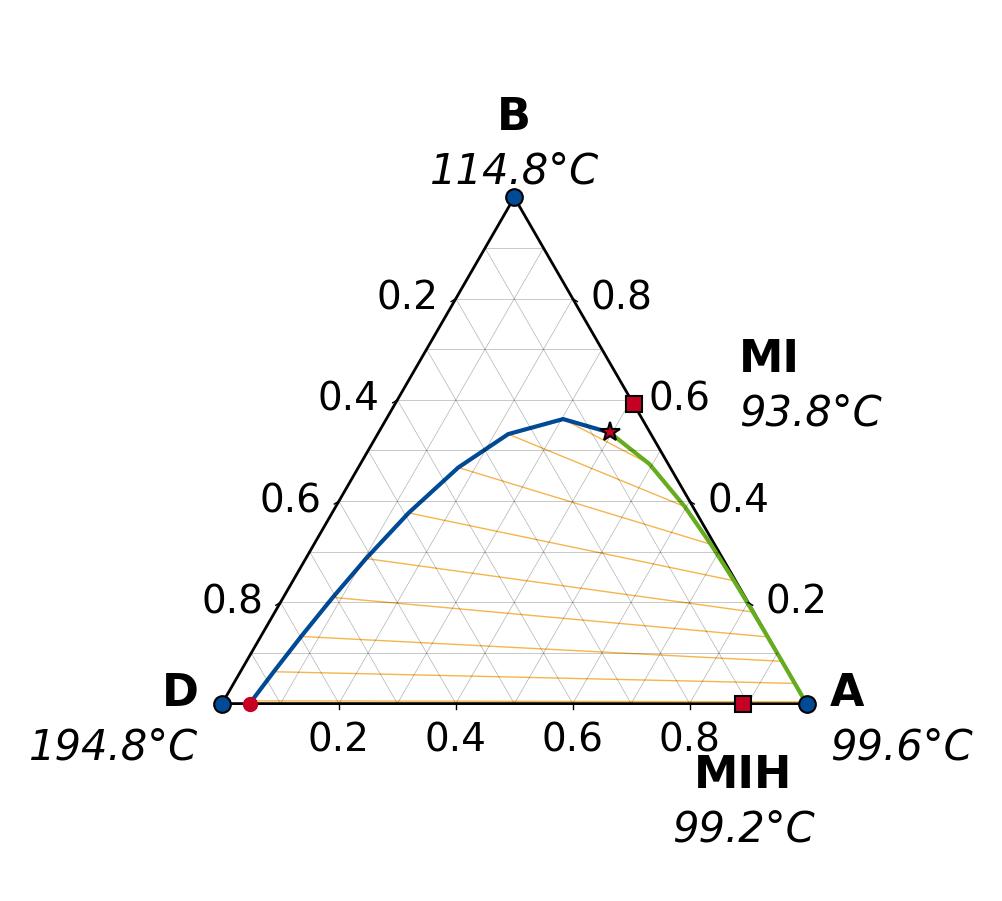}
		\caption{Entrainer candidate D being a high-boiler, preventing B recovery in heteroazeotropic distillation.}
		\label{fig:tern_pyr_oct}
	\end{subfigure}
	\hfill
	\begin{subfigure}[b]{0.32\textwidth}
		\centering
		\includegraphics[width=\textwidth]{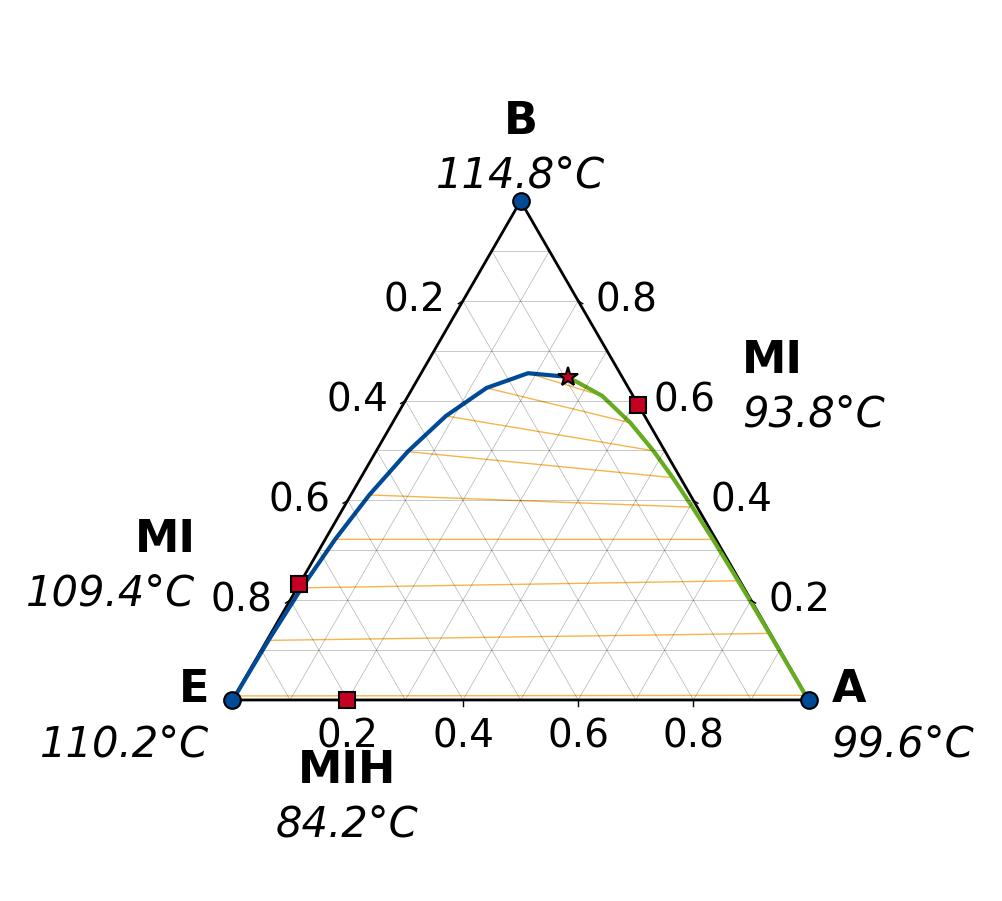}
		\caption{Entrainer candidate E forming a second homogeneous minimum azeotrope with B.}
		\label{fig:tern_pyr_tol}
	\end{subfigure}
	\caption{Ternary diagrams for case study 4 (more challenging scenarios for heteroazeotropic distillation and entrainer selection). Diagrams include the azeotropes and (ternary) miscibility gaps. Squares mark azeotropes, with MI denoting homogeneous minimum azeotropes and MIH heterogeneous minimum azeotropes. Thin lines inside the miscibility gap show tie lines connecting liquid phases in equilibrium. The critical point of the liquid-liquid equilibrium is marked with a star.}
	\label{fig:tern_pyr}
\end{figure}

Consider, for instance, case study 4 with a ternary system of water (A), pyridine (B), and entrainer candidates for heteroazeotropic distillation dimethyl carbonate (C), n-octanol (D), and toluene (E) (cf. Figure~\ref{fig:tern_pyr}).
 In this case, the literature considers toluene as an entrainer for separating A from B \citep{Chien2004,Gottl2025} due to a favorable miscibility gap. On a conceptual level, a process similar to the one depicted in Figure~\ref{fig:pfd_case3a} would be possible.
 However, it will be very challenging from a numerical perspective as the two curved distillation borders (from the homogeneous A-B (MI) and B-C (MI) azeotropes, respectively, to the heterogeneous A-C azeotrope (MIH)) almost touch each other.
 Based solely on the analysis of the miscibility gaps, candidate D would be an appealing entrainer as well; however, its high boiling point would prevent the recovery of B as bottom product in the distillation column, thus prohibiting a process concept as in Figure~\ref{fig:pfd_case3a}.
 Consequently, candidate C would be the most viable entrainer selection and is always chosen by the agent even though the partial mutual miscibility of A and C would require additional purification steps to obtain the required high purity of the A-rich phase from the decanter.
 Interestingly, if prohibiting the agent from using candidate C, it selects candidate D and at some point always ends in confusion:
 That is, the agent tries to design a sequence first using an extraction column to withdraw B from the feed using D as solvent. Subsequently, one could separate the extract containing mostly D and B with some minor A concentration by means of distillation. Note that this concept would indeed work and could be planned by the multi-agent system if instructed precisely and subsequently simulated.
 However, the agent at some point always switches its plan to an extractive distillation sequence instead. Here, D shall serve as entrainer, which is not a viable design and contradicts common guidelines for entrainer selection for extractive distillation (see, e.g., \cite{Kossack2008}).
 In contrast, a two-column process with E as entrainer would be a viable design but is never proposed by the agent who either considers the third azeotrope as a show-stopper or entirely ignores it.

\begin{figure}[htbp]
	\centering
	\begin{subfigure}[b]{0.48\textwidth}
		\centering
		\includegraphics[width=\textwidth]{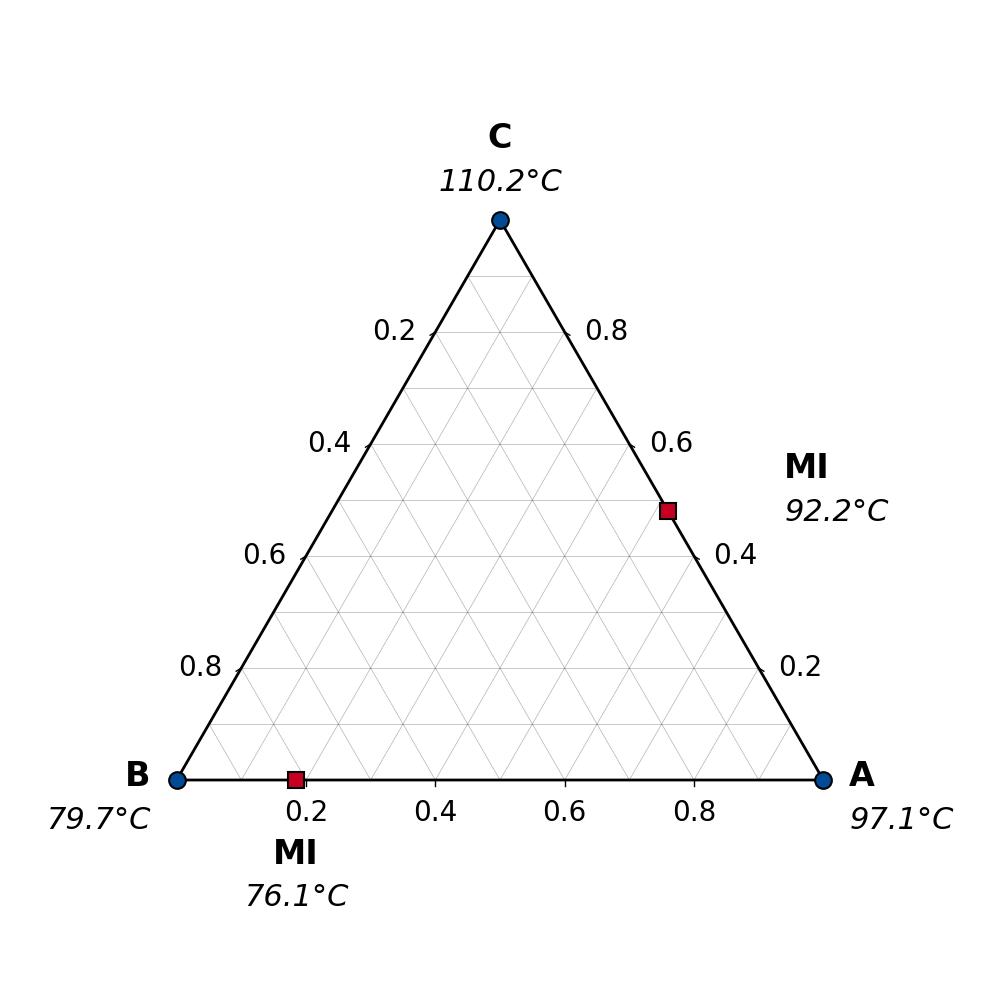}
		\caption{Low-pressure ternary diagram.}
		\label{fig:tern_lp}
	\end{subfigure}
	\hfill
	\begin{subfigure}[b]{0.48\textwidth}
		\centering
		\includegraphics[width=\textwidth]{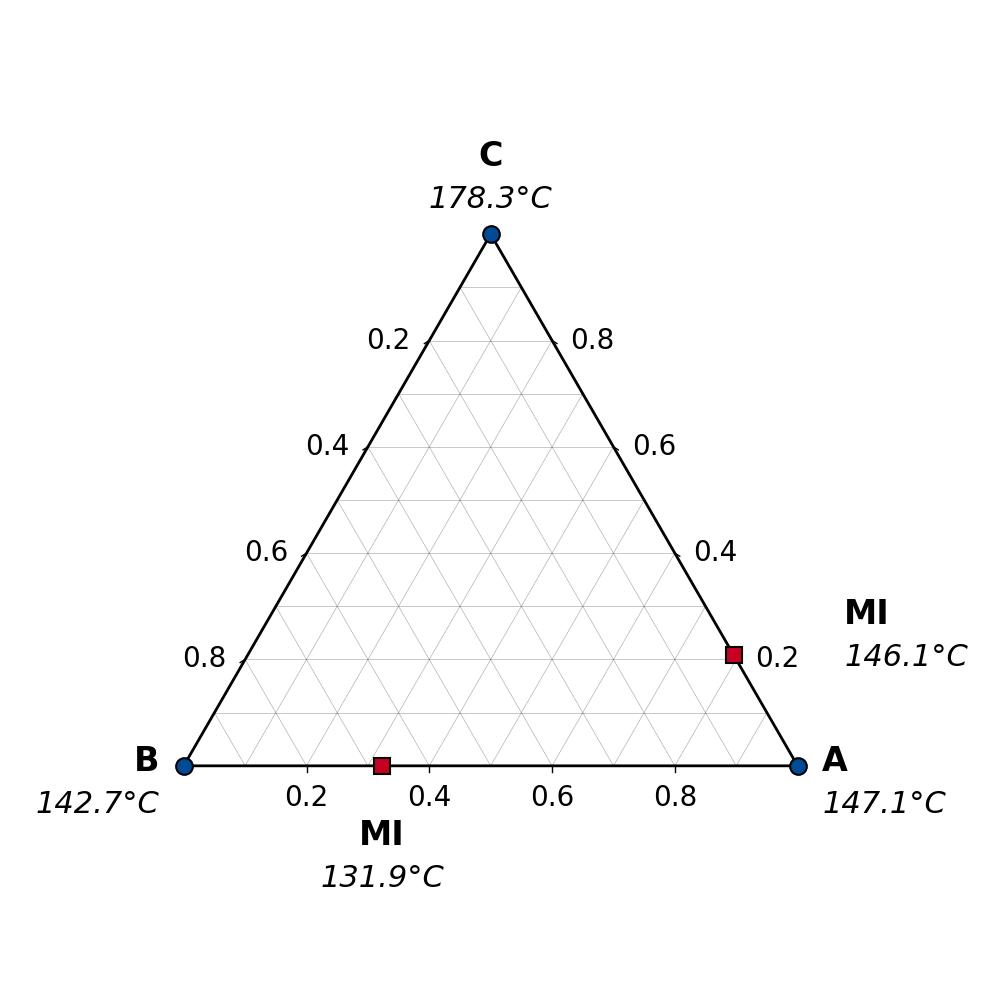}
		\caption{High-pressure ternary diagram.}
		\label{fig:tern_hp}
	\end{subfigure}
	\caption{Diagrams for case study 5 (ternary pressure-swing distillation). Diagrams show the pressure-dependence of the binary azeotropes. Squares mark azeotropes, with MI denoting minimum azeotropes.}
	\label{fig:tern_psd}
\end{figure}

Along the same lines, the multi-agent system fails to suggest a process design for separating a ternary mixture of n-propanol (A), benzene (B) and toluene (C) in case study 5 (cf. Figure~\ref{fig:tern_psd}) with the feed being located in the distillation region defined by the following nodes: pure A, homogeneous A-B azeotrope (MI) and homogeneous A-C minimum azeotrope (MI).
 In that case, a procedure for crossing the distillation border between the two regions which connects the two azeotropes would be feasible. Note that this is highly comparable to a heteroazeotropic distillation sequence as presented in case study 3.
More precisely, a third component (which is in this case already present in the feed) is used to shift the feed composition to the other distillation region where full withdrawal of one pure component is possible.
 That is, adding an excess of pure B (that can later be recycled from the product stream) would thus allow for the withdrawal of pure C as bottom product in the first column. This leaves a C-free A-B mixture as top product, which can be separated in two further distillation columns by means of a pressure-swing distillation sequence (as described above).
 However, even with a hint to the agent to consider recycling of pure components, it fails to design such a process.
 Instead, it suggests adding an excess of C, which also allows for crossing the distillation border. The plan to subsequently withdraw pure C as bottom product in the first column and achieve a C-free mixture does, however, not make sense (in that case, the top product would obviously be the intersection of the straight line between the feed and the pure C node with the distillation border).

As stated above, examples like these are already quite challenging and go beyond most of the common case studies in the literature.
 It is thus not fully surprising that the agentic AI system is currently not able to cope with designing processes for such systems.
 At the same time, we cannot expect that future generations of LLMs will be able to cope with even more complex systems.
 Likewise, there will not be much context available for the agent to learn from in these cases.
 Consequently, future work should focus on improving the capabilities of the agentic AI system to analyze complex thermodynamic behavior and design appropriate processes.
 This could be achieved by providing the agent with more specific tools for analyzing the thermodynamic behavior.
 In case of distillation processes as example, these could aim at identifying the distillation regions for arbitrary systems (e.g., \cite{Rooks1998}) and numerically checking feasibility of the desired separation sequences (e.g., \cite{Bausa1998,Watzdorf1999,Ryll2012}).
 Along these lines, the agent could also be provided with tools for performing sophisticated short-cut calculations for other separation processes, e.g., multi-stage extraction (e.g., \cite{Redepenning2017}).
 Ideally, enabling agents with the capability to use tools like these should be combined with the build-up (and maintenance) of structured knowledge databases (cf. \cite{Prasopsanti2026}) to further enhance the reasoning capabilities of the agentic AI system in terms of abstract process design tasks.

\paragraph{Equipment sizing and full flowsheet optimization}

In its current state, the multi-agent system suggests a process design and does not target any optimization except for a very basic reasoning about the performance (cf. case study 3).
 This limitation can be tackled on multiple levels. First, and likely most intuitively, the multi-agent system can be granted access to numerical tools for process optimization inside the process simulation environment itself, which in case of Chemasim corresponds to using a (mixed-integer) sequential quadratic programming method \citep{Exler2007} in a sequential (shooting) manner for identifying optimal process parameters.
 In that case, one could address the quantitative economic performance indicators, such as costs or profit.

Second, opportunities for a more quantitative assessment and optimization should be made available also at the conceptual design level (here, the process development agent).
 For instance, the multi-agent system could benefit from short-cut models for estimating the energy demand of the process (e.g., \cite{Bausa1998}) beyond solely solving the mass balance.
 Along these lines, there should also be a more explicit consideration of integrated or intensified process designs - also before the rigorous simulation step - to further enhance the economic performance.
 Here, the latter might be addressed in a promising way by providing additional context to the agent and more explicitly instructing it to consider improvement opportunities in the process design, e.g., by instructing the agent to consider opportunities for heat integration.

Third and in particular for more complex integrated or intensified processes, the equipment (pre-)sizing before the rigorous simulation step might become more important.
 For the considered case studies, the estimates of stages and reflux ratio for the distillation columns were in fact quite good and could be corrected in a trial-and-error manner by the Chemasim modelling agent to avoid over-separation and a massive waste of energy.
 Consequently, using Chemasim's capabilities for numerical optimization of the process parameters would have likely been sufficient to further optimize the design.
 If going, for instance, towards dividing wall columns to lower energy demand, tools could be used in order to ensure a good initial design \citep{Halvorsen2003}.

\paragraph{Interpretation of simulation results and convergence issues}

The capabilities of the Chemasim modelling agent for proper handling of the process simulation engine have not been limiting in this work.
 In fact, together with the best-practices provided within the system prompt, the accurate estimates of the process development agent for the mass balance and thus of the recycle streams ensured robust convergence of the simulations.
 This still holds if implementing the more complex process designs for the limiting case studies in Figures~\ref{fig:tern_pyr} and \ref{fig:tern_psd} as long as they are properly planned by the process development agent under our guidance.

If, however, facing the issue that a process is not properly planned, e.g., a mass balance does not close or - as happened for the ternary pressure-swing distillation - an intended separation is simply not feasible, the Chemasim modelling agent is currently not able to handle these issues.
 Here, we see a great potential in using the feedback opportunity from one agent to the other, which is technically already implemented but remains mostly unused in the current work.
 We believe that this will rather be an issue of appropriate system prompts and orchestration of the agents.

\section{Conclusion}

We present a multi-agent system for autonomous process development based on the interaction of two custom agents with distinct responsibilities.
 Therein, the process development agent solves abstract process synthesis tasks by analyzing the thermodynamic system using dedicated tools and performing mass balance computations via free coding capabilities, whereas the Chemasim modelling agent implements the resulting process design in valid Chemasim syntax and handles the simulation engine.
 Being text-based, BASF's in-house process modelling tool Chemasim is well suited for this approach as it allows the agent to directly manipulate the input files with the same flexibility as a human user.

We demonstrated the effectiveness of the framework on three case studies covering a reaction-separation process, pressure-swing distillations for binary azeotropes, and heteroazeotropic distillations with entrainer selection.
 In all cases, the process development agent is able to design reasonable process flowsheets based solely on the analysis of the thermodynamic behavior and its inherent process engineering knowledge - without any specific instructions on how to solve the respective synthesis problem.
 The Chemasim modelling agent subsequently implements these designs following a unit-by-unit procedure, guided by best-practices for ensuring convergent simulations. The obtained rigorous simulation results are in very good agreement with the simplified mass balance estimates of the process development agent.

At the same time, the case studies reveal current limitations of the agentic AI system, in particular regarding the capabilities for analysis and interpretation of complex thermodynamic behavior going beyond the common case studies in the literature.
 Here, we identified the provision of more specific tools for analyzing thermodynamic systems, checking feasibility of separation sequences and performing short-cut calculations as the most promising directions for future work.
 Furthermore, the integration of numerical optimization and a more systematic interaction between the agents could significantly enhance the capabilities of the framework.

We believe that the ongoing advancement of LLMs combined with the proposed extensions of the framework will allow for valuable decision support for chemical process development in the future.
 From the software development perspective, we expect a decreasing importance of graphical user interfaces and a shift towards text-based interaction with expert tools such as process flowsheet simulators.

\section*{Acknowledgments}
We thank Prof. Alexander Mitsos (RWTH Aachen University) for making Lukas Krinke's internship at BASF possible, during which we laid the foundation for the presented work.

\section*{Declaration of AI Assistance}
The authors confirm that the research work, scientific analysis, and the main intellectual contributions reported in this manuscript were conceived, conducted, and interpreted by the authors. Generative AI tools were used as supportive aids.
Specifically, AI assistance was used for (i) proofreading (spelling and grammar) and wording improvements during manuscript preparation and (ii) coding assistance during the research workflow, including support in writing scripts to use thermodynamic analysis tools and in formulating system prompts and task descriptions for the agentic framework in a concise and clear manner.
All AI-assisted outputs were reviewed and edited by the authors, who take full responsibility for the content of the manuscript.

\bibliographystyle{plainnat}
\bibliography{references}

\end{document}